\documentclass[lettersize,journal]{IEEEtran}
\usepackage{amsmath,amsfonts}
\usepackage{amsmath,amssymb,amsfonts,amsthm}
\usepackage{graphicx}
\usepackage{booktabs}
\usepackage{makecell}
\usepackage{multirow}
\usepackage{colortbl}
\usepackage{xcolor}
\usepackage{microtype}
\usepackage{xspace}

\usepackage{nicefrac}
\usepackage{algorithm}
\usepackage{algorithmic}
\usepackage{wrapfig}
\usepackage[caption=false,font=normalsize,labelfont=sf,textfont=sf]{subfig}
\usepackage{float}
\usepackage{enumitem}
\usepackage{tikz}
\usepackage{soul}
\usepackage{bbm}
\usepackage{placeins}
\usepackage{xurl}
\usepackage{url}
\definecolor{Red}{rgb}{0.6,0,0}
\definecolor{Blue}{rgb}{0,0,0.8}
\definecolor{Green}{rgb}{0.0,0.55,0.0}
\definecolor{mountainmeadow}{rgb}{0.19, 0.73, 0.56}
\definecolor{crimson}{rgb}{0.86, 0.08, 0.24}
\definecolor{darkblue}{rgb}{0.0, 0.0, 0.55}
\definecolor{ggr}{gray}{0.92}
\definecolor{gg}{HTML}{E0FEFE}
\newcolumntype{a}{>{\columncolor{ggr}}c}
\newtheorem{theorem}{Theorem}

\newtheorem{definition}{Definition}

\newcommand{\E}{\mathbb{E}}

\newcommand{\doop}{\mathrm{do}}
\newcommand{\CTIS}{\mathrm{CTIS}}

\newcommand{\GS}{\mathrm{GradScore}}

\usepackage[style=ieee,backend=bibtex]{biblatex}
\usepackage[
pdfauthor={derajan},
pdftitle={How to do this},
pdfstartview=XYZ,
bookmarks=true,
colorlinks=true,
linkcolor=blue,
urlcolor=blue,
citecolor=blue,
pdftex,
linktocpage=true,   
hyperindex=true
]{hyperref}
\makeatother
\begin{document}

\title{Beyond Correlation: Causality-Driven Visual Token Reduction for MLLMs via Counterfactual Intervention}

\author{Pengcheng Zheng, Chaoning Zhang, \IEEEmembership{Senior Member, IEEE}, Ya Wen, Jiaxin Huang, Jiaquan Zhang, Caiyan Qin, Kuien Liu, Sheng Zheng, Pietro Lió, \IEEEmembership{Fellow, IEEE}, and Yang Yang, \IEEEmembership{Senior Member, IEEE}
\thanks{This work was supported by the National Natural Science Foundation of China under grant 62572104 and 62220106008.} 
\thanks{Pengcheng Zheng, Chaoning Zhang, Ya Wen, Jiaquan Zhang, Sheng Zheng, and Yang Yang are with the Center for Future Media and School
of Computer Science and Engineering, University of Electronic Science and Technology of China, Chengdu 611731, China (e-mail: zpc777@std.uestc.edu.cn; chaoningzhang1990@gmail.com).}
\thanks{Jiaxin Huang is with the Department of Machine Learning, Mohamed bin Zayed University of Artificial Intelligence, Abu Dhabi, United Arab Emirates.}
\thanks{Caiyan Qin is with the School of Robotics and Advanced Manufacture, Harbin Institute of Technology, Shenzhen 518055, China.}
\thanks{Kuien Liu is with the Institute of Software, Chinese Academy of Sciences, Beijing 100190, China}
\thanks{Pietro Lió is with the Department of Computer Science and Technology, University of Cambridge, GB-CAM Cambridge, U.K.}
}

\markboth{Journal of \LaTeX\ Class Files,~Vol.~14, No.~8, August~2021}%
{Shell \MakeLowercase{\textit{et al.}}: A Sample Article Using IEEEtran.cls for IEEE Journals}


\makeatletter
\let\savedmaketitle\maketitle
\let\saved@maketitle\@maketitle
\let\savedthanks\thanks
\makeatother

\maketitle

\begin{abstract}
Multimodal Large Language Models (MLLMs) have achieved remarkable progress in multimodal understanding, but their practical deployment is still limited by the substantial computation introduced by excessive visual tokens. Visual token reduction offers an effective solution by removing redundant visual tokens during inference. However, attention- and feature-based correlation scores need not reflect how strongly a fixed model responds to perturbing each token. 
To address this issue, we propose \textbf{CAPruner}, a causality-driven visual token reduction framework that formulates token importance in terms of controlled model-internal interventions. We describe MLLM inference through a computational intervention model and introduce the Causal Token Importance Score (CTIS) as the absolute change in a scalar model response induced by zeroing an individual visual token at its original position while keeping other inputs fixed. To make this sensitivity-guided selection practical, we formulate an early-readout gradient-based importance proxy and refine it through a calibrated visual–text association correction. The resulting score is complemented by one-layer-lookahead hidden-state sensitivity estimates, enabling progressive pruning across transformer layers. Experiments across various MLLM architectures, such as LLaVA-1.5 and Qwen2.5-VL, show favorable performance retention under visual-token compression, together with reduced end-to-end inference latency compared with the unpruned backbone.
\end{abstract}


\section{Introduction}
\begin{figure}[!t] 
    \centering
\includegraphics[width=\linewidth]{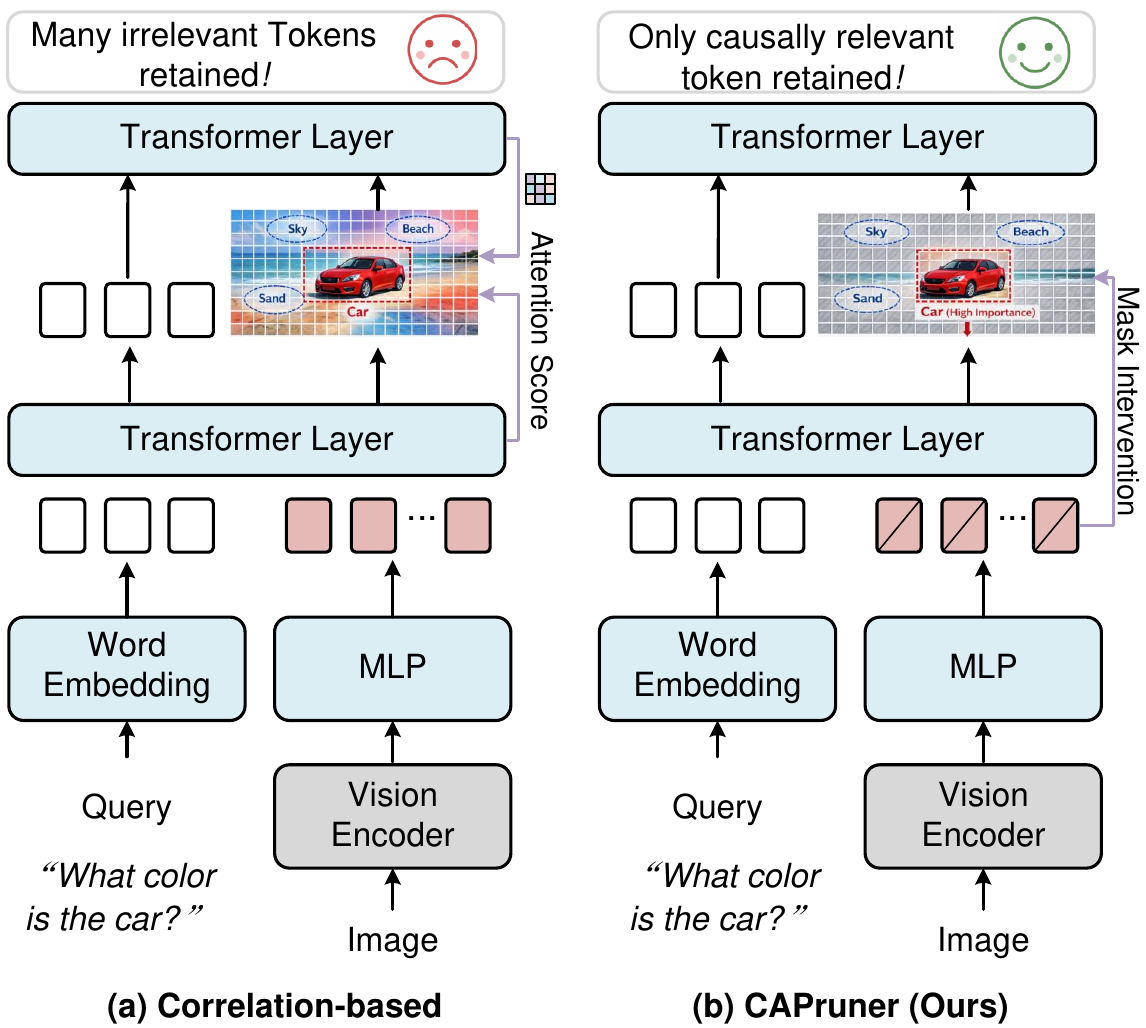}
\caption{
Motivation for CAPruner. \textbf{Left:} A proxy score emphasizes contextual tokens (beach, sky, sand). \textbf{Right:} A token-response-oriented score emphasizes the car region. 
}
\label{fig:motivation}
\end{figure}
Recently, Multimodal Large Language Models (MLLMs), such as LLaVA~\cite{liu2023visual}, InternVL~\cite{chen2024internvl}, and Qwen-VL~\cite{bai2023versatile}, have achieved remarkable progress across a broad spectrum of multimodal tasks, including visual question answering~\cite{wang2026towards,lu2025mitigating,huang2026metaphorical}, image captioning~\cite{zhu2023prompt,song2026image,wang2025seacap,zheng2026joint}, and visual reasoning~\cite{xiang2026atomthink,xu2026mmtot,li2026visual,zhou2026injepa}. These models typically extract visual representations with a visual encoder, integrate them with textual embeddings, and process the resulting multimodal sequence using Large Language Models (LLMs), thereby enabling strong cross-modal understanding and reasoning. Despite their impressive capabilities, practical deployment remains constrained by the substantial memory footprint and computational overhead caused by processing excessive visual tokens~\cite{cao2026madtp++,zhang2024sparsevlm}, especially in resource-limited environments.

To address this challenge, prior works~\cite{cao2026language,DART,visionzip,li2026hero,yan2026remic,zhang2026rcp} compress visual representations by pruning redundant tokens, motivated by the higher spatial redundancy of visual information than natural language~\cite{mao2025prune,jiang2025rethinking}. Early studies on Vision Transformers, such as DynamicViT~\cite{rao2021dynamicvit}, demonstrate that dynamically removing less informative visual tokens can significantly reduce computational overhead while preserving recognition performance. Building upon this idea, recent MLLM pruning methods further explore token reduction under multimodal reasoning scenarios. Existing methods mainly fall into three paradigms: attention-based methods, \textit{e.g.,} FastV~\cite{FastV}, SparseVLM~\cite{zhang2024sparsevlm}, and PyramidDrop~\cite{pyramiddrop}, score tokens by cross-modal attention under the assumption that highly attended tokens are more informative. Gradient-based methods, including ZOO-Prune~\cite{ZOOPrune} and SDTP~\cite{SDTP}, use loss gradients with respect to token embeddings to estimate each token's influence on the output. Diversity-based methods, \textit{e.g.,} DivPrune~\cite{divprune} and CDPruner~\cite{CDPruner}, retain diverse token subsets to maximize information coverage while removing redundant tokens.

Despite their success, most existing token-selection scores are not explicitly defined against a fixed-target final-model response under controlled token masking. As shown in Figure~\ref{fig:motivation}, for the query ``What color is the car?'' a pruning score may assign substantial weight to beach or sky regions, even though a focused answer should depend on the car's appearance. 
\textit{Pruning methods based only on proxy importance signals may retain less useful tokens while discarding tokens whose removal substantially changes a prediction.} Motivated by this observation, we ask a natural question: \textit{how can we estimate the effect of perturbing each visual token on a fixed MLLM's output without exhaustively rerunning the model?}
A useful framework for addressing this question is intervention-based analysis, which studies how an output changes when a specified input variable is perturbed~\cite{neuberg2003causality,pearl2009causal}. For a frozen MLLM, a controlled token replacement defines a model-internal intervention. Its effect on the model output can be evaluated without identifying population-level causal effects from observational data. This perspective is related to counterfactual perturbation studies in vision-language modeling~\cite{li2025dcdl,wang2025causalrag,zhang2025cf,gao2025causality,li2025treble}. 
At the token level, although gradient- and perturbation-based importance estimation has been explored for MLLM pruning, explicitly characterizing token importance through controlled interventions on the final model response remains relatively underexplored.
This motivates a shift from token relevance toward model-output influence, understood as the response to controlled token perturbations. This raises an intuitive question: \textit{can a model-response-oriented token reduction strategy complement attention-based selection in MLLMs?} Addressing this question requires characterizing token-level perturbation sensitivity while maintaining computational tractability. To this end, we propose \textbf{CAPruner}, based on the key insight that the question is not only \textit{which tokens does the model attend to?} but also \textit{which tokens substantially affect its output when perturbed?}
We describe MLLM inference using a structural computation model and define the Causal Token Importance Score (CTIS) as the absolute change in a fixed scalar model response under a controlled zero-reference token intervention. The key distinction lies in using this intervention response as an explicit reference for guiding token importance estimation. To avoid exhaustive token-wise interventions, we formulate an early-readout gradient-based importance proxy and refine it through calibrated visual--text association statistics. Building on this formulation, we exploit one-layer-lookahead hidden-state sensitivity estimates for progressive pruning, allowing token importance to be updated across transformer depth.

The main contributions of this paper are as follows:
\begin{itemize}[leftmargin=*, itemsep=0.3ex, parsep=0ex, topsep=0.3ex]
\item We present CAPruner, a causality-driven framework for visual token reduction in MLLMs that defines token importance through the change in a fixed-target model response under controlled token masking.
\item We formulate an early-readout gradient-based sensitivity proxy for CTIS and refine the resulting token scores with offline-calibrated visual--text association statistics.
\item We propose a one-layer-lookahead hidden-state sensitivity formulation for progressive visual token reduction, enabling token rankings to be updated across transformer layers without repeated full-model backpropagation.
\item Extensive experiments across MLLMs and benchmarks demonstrate favorable performance retention at pruning ratios up to 94.4\%, together with reduced end-to-end inference latency relative to the unpruned backbone.
\end{itemize}

\section{Related Work}
\label{headings}

\subsection{Visual Token Reduction in MLLMs}
Visual token reduction in MLLMs has evolved from early heuristics to adaptive, query-aware selection. FastV~\cite{FastV} exploits early-layer features for coarse-to-fine pruning, reducing computation but risking the removal of semantically important tokens. Building on adaptive selection, ATP-LLaVA~\cite{ye2025atp} ranks visual tokens conditioned on textual queries via cross-modal attention to improve redundancy reduction, while SparseVLM~\cite{zhang2024sparsevlm} estimates token saliency from cross-modal interactions to remove redundant tokens, yet still relies on observational relevance signals. CDPruner~\cite{CDPruner} further preserves representative and diverse visual tokens for aggressive compression, while remaining dependent on proxy selection criteria.
Recent approaches further explore sensitivity- and
transition-based pruning. ZOO-Prune~\cite{ZOOPrune}
estimates token sensitivity through zeroth-order
perturbations at the projection layer and combines
sensitivity with feature diversity, while
TransPrune~\cite{li2025transprune} exploits token transition
variation and instruction-guided attention.
Unlike projection-level sensitivity estimation,
CAPruner uses the fixed-target prediction response
of the final MLLM under controlled token masking
as its intervention reference. Rather than introducing
a new gradient attribution operator, CAPruner develops
an early-readout sensitivity proxy with cross-modal
association correction and depth-dependent sensitivity
updates for progressive token reduction.
\subsection{Causal Inference in Vision and Language}
Causal inference has been increasingly explored in vision-language modeling, spanning representation-level bias mitigation to fine-grained token-level attribution~\cite{neuberg2003causality,scholkopf2021toward}. CausalVAE~\cite{yang2021causalvae} learns disentangled visual representations by modeling latent causal factors with structural causal models, mitigating spurious correlations while remaining agnostic to downstream token-level contributions. CausalRAG~\cite{wang2025causalrag} adjusts causal effects between retrieved evidence and generated outputs to improve representation-level alignment. In parallel, CF-VLM~\cite{zhang2025cf} estimates token-level influence via counterfactual perturbations of visual inputs and their effects on generated captions, enabling post hoc attribution. More recently, CausalAgent~\cite{zhu2026causalagent} uses intervention-based causal modeling to disentangle spurious correlations between visual features and semantic concepts, reducing representation bias while remaining agnostic to token-level contributions. 
In this work, we adopt a model-internal intervention perspective for visual token reduction in MLLMs, defining token importance by the model-output change induced by controlled token masking.

\section{Methodology}
\label{others}
The MLLM architecture and the formal definition of the visual
token reduction problem are both provided in the accompanying
\textcolor{red}{Supplementary Material}.
\subsection{Preliminary Knowledge on Causal Inference}
In this section, we introduce the key concepts of causal inference, including SCMs, interventions, ATE, and the backdoor criterion, which provide the necessary theoretical foundation for our proposed CAPruner framework.

\begin{definition}[Structural Causal Model]
\label{def:scm}
An SCM $\mathcal{M} = (\mathbf{U}, \mathbf{X}, \mathcal{F}, P(\mathbf{U}))$
consists of: a set of exogenous (background) variables $\mathbf{U}$ with joint
distribution $P(\mathbf{U})$; a set of endogenous variables $\mathbf{X}$; and a set of
structural equations $\mathcal{F} = \{f_i\}$ where each $X_i = f_i(\mathrm{Pa}(X_i), U_i)$
expresses $X_i$ as a deterministic function of its causal parents $\mathrm{Pa}(X_i)$ and
its exogenous noise $U_i$. The SCM induces a directed acyclic graph (DAG) $\mathcal{G}$
where edges represent direct causal relationships.
\end{definition}

\begin{definition}[Do-operator and Average Treatment Effect]
\label{def:ate}
The intervention $\doop(X_i = x)$ in SCM $\mathcal{M}$ sets $X_i$ to value $x$ by
replacing its structural equation $f_i$ with the constant $x$, while leaving all other
equations unchanged. The resulting interventional distribution is denoted
$P(Y \mid \doop(X_i = x))$. The ATE of $X_i$ comparing
treatment $x_i$ to control $x_0$ is:
\begin{equation}
\small
\mathrm{ATE}(X_i;\, x_i, x_0) = \E[Y \mid \doop(X_i = x_i)] - \E[Y \mid \doop(X_i = x_0)].
\end{equation}
\end{definition}

\begin{definition}[Backdoor Criterion and Adjustment]
\label{def:backdoor}
A set of variables $\mathbf{Z}$ satisfies the backdoor criterion relative to the ordered
pair $(X_i, Y)$ in DAG $\mathcal{G}$ if: (1) no node in $\mathbf{Z}$ is a descendant of
$X_i$, and (2) $\mathbf{Z}$ blocks every path between $X_i$ and $Y$ that contains an
arrow into $X_i$ (a ``backdoor path''). When $\mathbf{Z}$ satisfies the backdoor
criterion, the interventional distribution is identifiable from observational data:
\begin{equation}
\small
P(Y \mid \doop(X_i = x)) = \sum_z P(Y \mid X_i = x, \mathbf{Z} = z)\, P(\mathbf{Z} = z).
\end{equation}
\end{definition}
\noindent This expression indicates that, \emph{only when a valid adjustment set exists}, a population causal effect can be identified from observational data. The present token-scoring method does not establish that text tokens satisfy this criterion, and does not use this formula to identify a population ATE.

\subsection{Causal Formulation of MLLM Inference}
\label{sec:causal_formulation}

\begin{figure}[t]
\centering
\includegraphics[width=\linewidth]{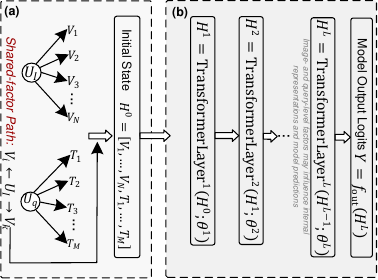}
\caption{Schematic computation graph of MLLM inference and potential shared image-level factors. 
}
\label{fig:Graph}
\end{figure}

We model a frozen MLLM as a structural computation system whose processing graph
$\mathcal{G}_{\text{MLLM}}$ is illustrated schematically in Figure~\ref{fig:Graph}. The variables are the visual tokens $V_i$, text tokens $T_j$, hidden states $H^l$, and output logits $Y$. The original structural equations are retained:
\begin{align}
V_i &= f_V(I, U_I)_i, \qquad T_j = f_T(q, U_q)_j \label{eq:scm_input}\\
H^0 &= [V_1, \ldots, V_N,\, T_1, \ldots, T_M] \label{eq:scm_h0}\\
H^l &= \mathrm{TransformerLayer}^l(H^{l-1};\, \theta^l), \quad l=1,\ldots,L \label{eq:scm_hl}\\
Y &= f_{\mathrm{out}}(H^L). \label{eq:scm_out}
\end{align}
Visual tokens can share image-level factors, so token dependencies and cross-token effects must be expected. In particular, the open path $V_i\leftarrow U_I\rightarrow V_k\rightarrow H\rightarrow Y$ is not blocked simply by conditioning on text tokens. We therefore use this graph to motivate controlled model perturbations, not to assert backdoor identifiability.
Let $\ell_0=\min\mathcal L_p$ denote the first pruning layer. We obtain an early prediction by applying the frozen LM head and output normalization to the last prompt-token representation $h_{\mathrm{ans}}^{\ell_0+1}$. Specifically, let $p_l(y)=\mathrm{softmax}(W_{\mathrm{LM}}\mathrm{Norm}(h_{\mathrm{ans}}^l))_y$, and select $y^*=\arg\max_y p_{\ell_0+1}(y)$ from the unpruned prefix. We define the final-model scalar response as $F(\mathbf V,\mathbf T)=\log p_\theta(y^*\mid\mathbf V,\mathbf T)$, with $y^*$ fixed across interventions. Let $F_{\doop(V_i=x)}$ denote this response when $V_i$ is set to $x$, while all other tokens remain unchanged. We define CTIS as
\begin{equation}
\CTIS(v_i)=
\left|
F_{\doop(V_i=v_i)}
-
F_{\doop(V_i=\mathbf{0})}
\right|.
\label{eq:ctis}
\end{equation}
For online scoring, we instead use the early scalar response $F_{\mathrm e}=\log p_{\ell_0+1}(y^*)$. At later pruning layers $l$, the local response is $F_{l+1}=\log p_{l+1}(y^*)$, evaluated on the currently retained sequence. These intermediate responses are computational proxies for the final-model response, not equivalent quantities.
This is a computational intervention on a fixed model, not a population ATE inferred from observational covariances. Masking at a fixed position also need not equal physically removing a token, which can change the sequence and attention normalization. 
As CTIS measures the magnitude of the response change, it does not distinguish whether a token positively or negatively contributes to the target output.
\begin{algorithm}[!t]
\caption{CAPruner: Sensitivity-Guided Visual Token Reduction}
\label{alg:causalpruning}

\begin{algorithmic}[1]
\REQUIRE Visual tokens $\mathbf V=\{v_i\}_{i=1}^N$,
text tokens $\mathbf T=\{t_j\}_{j=1}^M$,
retained-token ratio $\rho\in(0,1)$,
pruning layers $\mathcal L_p\subseteq[L-1]$,
association coefficients $\{\hat\gamma_{pq}\}$,
mixing coefficient $\alpha$
\ENSURE Retained token index set $\mathcal S$
with $|\mathcal S|=\lfloor\rho N\rfloor$

\STATE \textbf{// Stage 1: Early gradient scoring}
\STATE $\ell_0\leftarrow\min\mathcal L_p$
\STATE Forward to layer $\ell_0+1$;
cache $H^{\ell_0}$
\STATE Select $y^*\leftarrow\arg\max_y p_{\ell_0+1}(y)$
\STATE $F_{\mathrm e}\leftarrow\log p_{\ell_0+1}(y^*)$
\STATE Backpropagate $F_{\mathrm e}$ through the prefix
to obtain input-token gradients and
$\nabla_{H_i^{\ell_0}}F_{\mathrm e}$
\STATE $\GS(v_i)\leftarrow
\|\nabla_{v_i}F_{\mathrm e}\odot v_i\|_2$

\STATE \textbf{// Stage 2: Cross-modal association correction}
\STATE Compute pooled textual scores $\{\bar b_q\}$
using $\nabla_{t_j}F_{\mathrm e}$
\STATE $S_{\mathrm{assoc}}(v_i)\leftarrow
\GS(v_i)-\sum_{q=1}^{B_T}
\hat\gamma_{b_V(i),q}\bar b_q$

\STATE \textbf{// Stage 3: Progressive pruning}
\STATE $\mathcal S\leftarrow[N]$
\FOR{each $l\in\mathcal L_p$ in increasing order}
    \IF{$l>\ell_0$}
        \STATE Forward to $H^l$ with retained tokens
        \STATE Look ahead through layer $l+1$
        to compute $F_{l+1}=\log p_{l+1}(y^*)$
        \STATE Backpropagate only through layer $l+1$
        to obtain $\nabla_{H_i^l}F_{l+1}$
    \ENDIF
    \STATE $S_{\mathrm{layer}}^l(v_i)\leftarrow
    \|\nabla_{H_i^l}F_{l+1}\odot H_i^l\|_2$
    for $i\in\mathcal S$
    \STATE $\widetilde S^l(v_i)\leftarrow
    \alpha\mathcal N_{\mathcal S}(S_{\mathrm{assoc}}(v_i))
    +(1-\alpha)\mathcal N_{\mathcal S}
    (S_{\mathrm{layer}}^l(v_i))$
    \STATE $k_l\leftarrow\lfloor\rho N\rfloor$
    if $l$ is the final pruning layer;
    otherwise
    $k_l\leftarrow
    \lfloor\rho^{1/|\mathcal L_p|}|\mathcal S|\rfloor$
    \STATE $\mathcal S\leftarrow
    \mathrm{TopK}(
    \{\widetilde S^l(v_i)\}_{i\in\mathcal S},k_l)$
    \STATE Prune $H^l$ to retain $\mathcal S$
    \STATE Recompute layer $l+1$
    on the retained sequence
\ENDFOR
\STATE Continue through the remaining layers
\RETURN $\mathcal S$
\end{algorithmic}
\end{algorithm}
\begin{figure*}[t]
    \centering
    \includegraphics[width=\textwidth]{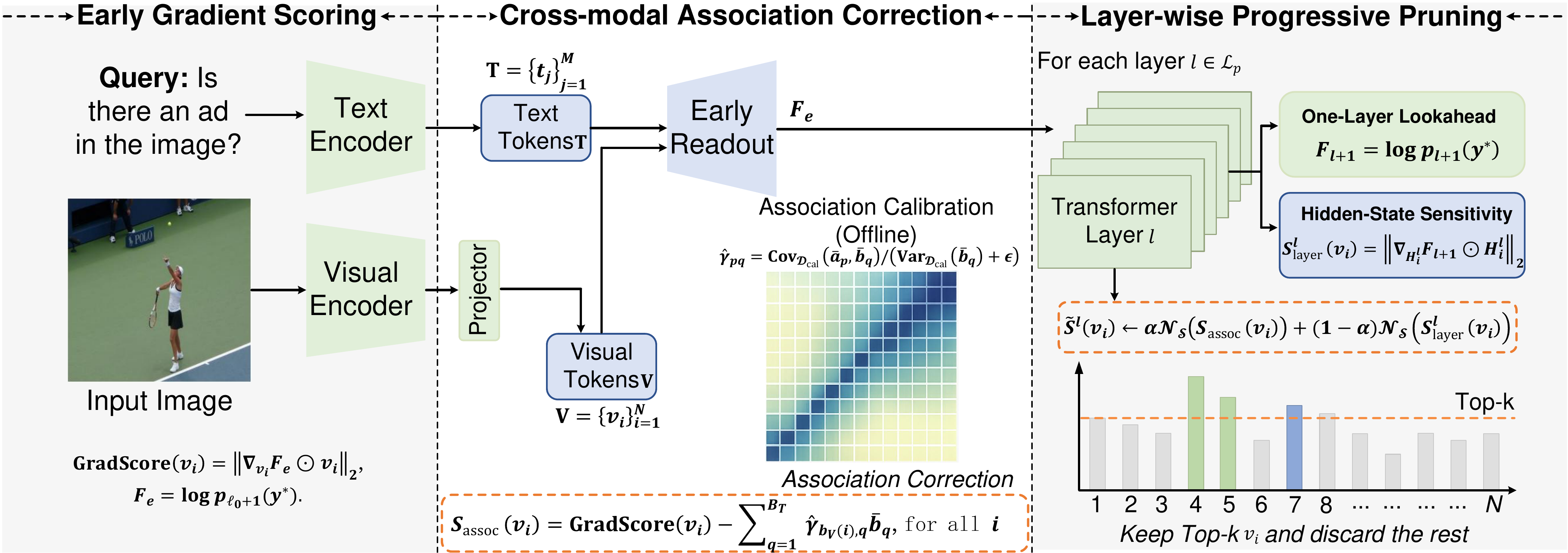}
\caption{
    Overall overview of CAPruner.
    CAPruner consists of three main stages: early gradient-based sensitivity scoring, cross-modal association correction using offline calibration statistics, and progressive visual token pruning guided by the corresponding one-layer-lookahead hidden-state sensitivity.
}
    \label{fig:efficiency}
    \vspace{-2mm}
\end{figure*}

\noindent\textbf{Discussion: Why Attention Weights Do Not Determine CTIS?}
Let $A_i^l = \frac{1}{M}\sum_{j=1}^M \alpha_{ij}^l$ denote the average attention weight
received by visual token $v_i$ at layer $l$ from text tokens, where
$\alpha_{ij}^l = \mathrm{softmax}(q_j^l \cdot k_i^l / \sqrt{d})_i$ is the attention
weight from text token $t_j$ to visual token $v_i$. This is the importance metric used
by FastV~\cite{FastV} and related methods.
\begin{theorem}[Attention Is Not Sufficient for Intervention Sensitivity]
\label{thm:bias}
For a fixed input containing a nonzero visual token $v_i\in\mathbb{R}^d$ with $d>1$, there exist two single-head attention networks $\theta_A$ and $\theta_B$ with identical attention weights but different scalar intervention sensitivities, i.e., $\CTIS_{\theta_A}(v_i)\neq\CTIS_{\theta_B}(v_i)$.
\end{theorem}

\begin{proof}
Let $\mathbf{x}=(x_1,\ldots,x_L)$ denote the input sequence, where $x_i=v_i\neq\mathbf{0}$. Consider a single-head attention network with scalar output
\begin{equation*}
z_{\theta_r}(\mathbf{x})
=w_r^\top\sum_{j=1}^{L}\alpha_jx_j,
\qquad
F_{\theta_r}(\mathbf{x})
=\log\sigma(c+z_{\theta_r}(\mathbf{x})),
\end{equation*}
where $r\in\{A,B\}$. 
Here $\sigma$ is the logistic sigmoid, and the shared bias $c$ is chosen sufficiently large so that the same target token is predicted by both unpruned networks. Setting all queries and keys to zero gives identical uniform attention weights $\alpha_j=1/L$ in both networks. For $d>1$, choose $w_A$ orthogonal to $v_i$ and let $w_B=v_i/\|v_i\|_2^2$. Under the intervention $v_i\leftarrow\mathbf{0}$, the attention weights remain unchanged. Therefore,
\begin{equation*}
\begin{aligned}
\CTIS_{\theta_r}(v_i)
=\Bigl|
&\log\sigma(c+z_{\theta_r}(\mathbf{x}))\\
-&\log\sigma(c+z_{\theta_r}(\mathbf{x})
-\alpha_i w_r^\top v_i)
\Bigr|.
\end{aligned}
\end{equation*}
which gives
\begin{equation*}
\CTIS_{\theta_A}(v_i)=0,
\qquad
\CTIS_{\theta_B}(v_i)>0.
\end{equation*}
Thus, identical attention weights can correspond to different intervention sensitivities.
\end{proof}
This result establishes the insufficiency of attention weights
alone for determining intervention sensitivity, motivating
response-based token scoring without implying its superiority
over attention-based alternatives.
\subsection{Gradient-Based Sensitivity Proxy}
\label{sec:efficient_estimation}

Exact computation of $\CTIS(v_i)$ by individual interventions can require one extra evaluation per visual token, which is impractical for large token sets. Therefore, we retain a gradient-based score as a tractable proxy, while distinguishing this score from the exact response difference.

\noindent\textbf{First-order Taylor Approximation.}
We approximate the early intervention response using
the differentiable scalar target $F_{\mathrm e}$
defined in Section~\ref{sec:causal_formulation}.
Along the zero-reference intervention path, Taylor's
theorem~\cite{nocedal2006numerical} gives
\begin{equation}
F_{\mathrm e}(\mathbf V)
-F_{\mathrm e}(\mathbf V^{(i\leftarrow0)})
=
\nabla_{v_i}F_{\mathrm e}(\mathbf V)^\top v_i+r_i,
\quad
|r_i|\leq
\frac{\Lambda_i}{2}\|v_i\|_2^2.
\label{eq:taylor}
\end{equation}
provided the Hessian spectral norm is at most $\Lambda_i$ along this intervention path. Following the original implementation, we use the element-wise gradient score
\begin{equation}
\GS(v_i)=
\left\|
\nabla_{v_i}F_{\mathrm e}\odot v_i
\right\|_2.
\label{eq:gradscore}
\end{equation}
The element-wise $\ell_2$ norm serves as a gradient-based
sensitivity proxy rather than an exact first-order
approximation of CTIS. Unlike the directional derivative
$|\nabla_{v_i} F_e^\top v_i|$, it preserves channel-wise
sensitivity magnitudes that may otherwise cancel.
While such cancellation-free aggregation does not
necessarily improve approximation accuracy, its empirical
ranking fidelity is examined in
Table~\ref{tab:early_scoring_fidelity}.

\noindent\textbf{Cross-modal Association Correction.}
We further introduce an empirical, text-conditioned
bin-level calibration to refine gradient-based token scores.
The motivation is that visual gradient scores may exhibit
systematic variations associated with textual sensitivities,
which need not directly reflect token-specific intervention
effects.
For each
calibration sample, we compute scalar visual and textual
sensitivity scores
$a_i=\|\nabla_{v_i}F_{\mathrm e}\odot v_i\|_2$
and
$b_j=\|\nabla_{t_j}F_{\mathrm e}\odot t_j\|_2$.
To accommodate varying token lengths, we partition visual
and textual tokens into $B_V$ and $B_T$ relative-position
bins, respectively, and denote their mean scores by
$\bar a_p$ and $\bar b_q$. The association coefficients
are estimated offline as
$\hat\gamma_{pq}=
\operatorname{Cov}_{\mathcal D_{\mathrm{cal}}}
(\bar a_p,\bar b_q)/
(\operatorname{Var}_{\mathcal D_{\mathrm{cal}}}
(\bar b_q)+\epsilon)$.
The corrected visual-token score is
\begin{equation}
S_{\mathrm{assoc}}(v_i)
=
\GS(v_i)-
\sum_{q=1}^{B_T}
\hat\gamma_{b_V(i),q}\bar b_q,
\label{eq:ctis_approx}
\end{equation}
where $b_V(i)$ denotes the corresponding visual bin containing $v_i$. 
The correction applies a text-conditioned offset shared
by all tokens within each visual bin, adjusting cross-bin
rankings while preserving within-bin ordering. The
pairwise coefficients provide a lightweight alternative
to joint multivariate regression, without explicitly
accounting for dependencies among textual bins.
Rather than assuming that visual--text associations are
necessarily spurious, this correction empirically
calibrates token rankings against intervention sensitivity,
without claiming causal identification.
\begin{theorem}[A Bound for the Early Gradient Proxy]
\label{thm:approx}
Let $\CTIS_{\mathrm e}(v_i)=
|F_{\mathrm e}(\mathbf V)
-F_{\mathrm e}(\mathbf V^{(i\leftarrow0)})|$
denote the early intervention response. Under the pathwise Hessian bound above and with visual embedding dimension $d$,
\begin{equation}
\CTIS_{\mathrm e}(v_i)
\leq
\sqrt{d}\,\GS(v_i)
+\frac{\Lambda_i}{2}\|v_i\|_2^2.
\label{eq:approx_bound}
\end{equation}
\end{theorem}
\noindent This bound establishes a theoretical connection between
GradScore and early intervention sensitivity, rather than
guaranteeing approximation accuracy or ranking consistency.
The result follows from Taylor's theorem and the inequality
$|\sum_k a_k| \leq \sqrt{d}\|\mathbf{a}\|_2$
for $\mathbf{a}=\nabla_{v_i}F_e\odot v_i$.
Its empirical alignment with final-model CTIS is evaluated
in Section~\ref{early}.
A detailed proof is provided in the \textcolor{red}{Supplementary Material}.

\subsection{Layer-wise Sensitivity for Progressive Pruning}
\label{sec:mediation}

A visual token may influence the model output not only through its own hidden state, but also through the hidden states of other tokens. At layer $l$, a first-order expansion of the output change gives
\begin{equation}
\Delta F_{l+1}\approx
\sum_k
\left(\nabla_{H_k^l}F_{l+1}\right)^\top
\Delta H_k^l.
\label{eq:tce}
\end{equation}
where the sum includes cross-token paths. This is a local chain-rule expansion, not an identification of natural direct and indirect effects. 
This expansion includes cross-token interactions. The layer-wise score below measures local hidden-state sensitivity rather than decomposing the total effect of an input-token intervention.

\begin{table*}[!t]
  \centering
\caption{
Main experimental results on Qwen2.5-VL-7B under different
visual token retention budgets. \textbf{Avg.} denotes the
average benchmark performance, and \textbf{Rel.} denotes
the mean of per-benchmark performance ratios relative to
the full-token baseline. The best results under each token
budget are highlighted in \textbf{bold}.
}
  \label{tab:qwen2.5-vl}
  \resizebox{\linewidth}{!}{
    \begin{tabular}{l|c|cccccccc|aa}
    \toprule
    \textbf{Method} &Venue & \textbf{TextVQA} & \textbf{ChartQA} & \textbf{AI2D} & \textbf{OCRBench} & \textbf{HallBench} & \textbf{MME} & \textbf{MMB-EN} & \textbf{MMB-CN} & \textbf{Avg.} & \textbf{Rel.} \\
    \rowcolor{ggr}\multicolumn{12}{c}{\textit{Full-token Baseline, All 1296 Tokens} ($\mathbf{100\%}$)} \\
    Qwen2.5-VL-7B &- & 84.8 & 86.1 & 80.4 & 863 & 46.8 & 2304 & 82.8 & 83.2 & 83.2 & 100.0\% \\
    \rowcolor{ggr}\multicolumn{12}{c}{\textit{Keep 512 Tokens} \textcolor{Green}{(\textbf{Pruning Ratio} $\mathbf{60.5\%}$)}} \\
    FastV &ECCV-2024 & 84.1 & 82.2 & 78.8 & 815 & 42.4 & 2317 & 82.0 & 81.8 & 81.1 &  97.0\% \\
    DivPrune &CVPR-2025 & 81.8 & 79.6 & 78.6 & 800 & \textbf{43.3} & 2279 & 81.6 & 82.1 & 80.1 &  96.0\% \\
    CDPruner &NeurIPS-2025 & 84.2 & 82.8 & 78.9 & 827 & 42.5 & 2327 & 82.2 & 82.6 & 81.5 & 97.5\% \\
\cellcolor{gg}\textbf{CAPruner}{ \small(Ours)}
& \cellcolor{gg}This paper
& \cellcolor{gg}\textbf{84.6}
& \cellcolor{gg}\textbf{83.4}
& \cellcolor{gg}\textbf{79.5}
& \cellcolor{gg}\textbf{830}
& \cellcolor{gg}43.2
& \cellcolor{gg}\textbf{2346}
& \cellcolor{gg}\textbf{82.8}
& \cellcolor{gg}\textbf{83.0}
& \textbf{82.1}
& \textbf{98.2\%}\\
    \rowcolor{ggr}\multicolumn{12}{c}{\textit{Keep 256 Tokens} \textcolor{Green}{(\textbf{Pruning Ratio} $\mathbf{80.2\%}$)}} \\
    FastV &ECCV-2024 & 81.5 & 70.9 & 76.2 & 703 & 39.0 & 2238 & 79.6 & 78.9 & 76.0 &  90.8\% \\
    DivPrune &CVPR-2025 & 76.0 & 65.1 & 76.5 & 692 & 36.4 & 2184 & 80.0 & 79.6 & 74.0 &  88.2\% \\
    CDPruner &NeurIPS-2025 & 82.4 & 73.0 & 77.5 & 749 & \textbf{40.1} & 2245 & 80.9 & \textbf{79.9} & 77.6 &  92.8\% \\
        
\cellcolor{gg}\textbf{CAPruner}{ \small(Ours)}
& \cellcolor{gg}This paper
& \cellcolor{gg}\textbf{84.4}
& \cellcolor{gg}\textbf{76.2}
& \cellcolor{gg}\textbf{78.3}
& \cellcolor{gg}\textbf{768}
& \cellcolor{gg}38.9
& \cellcolor{gg}\textbf{2314}
& \cellcolor{gg}\textbf{81.9}
& \cellcolor{gg}78.2
& \textbf{78.8}
& \textbf{93.9\%}\\
    \rowcolor{ggr}\multicolumn{12}{c}{\textit{Keep 128 Tokens} \textcolor{Green}{(\textbf{Pruning Ratio} $\mathbf{90.1\%}$)}} \\
    FastV &ECCV-2024 & 73.8 & 52.2 & 71.4 & 531 & 33.8 & 2008 & 72.9 & 72.2 & 66.2 &  79.0\% \\
    DivPrune &CVPR-2025 & 67.0 & 50.4 & 72.1 & 549 & 32.6 & 2108 & 77.8 & \textbf{77.8} & 67.3 &  79.9\% \\
    CDPruner &NeurIPS-2025 & 77.8 & 59.2 & 74.0 & 632 & 37.2 & 2127 & 76.2 & 76.5 & 71.3 &  85.2\% \\
    IDPruner &arXiv-2026 & 74.4 & 62.5 & 75.2 & 625 & 38.0 & 2114 & 74.7 & 71.7 & 70.6 &  84.5\% \\
\cellcolor{gg}\textbf{CAPruner}{ \small(Ours)}
& \cellcolor{gg}This paper
& \cellcolor{gg}\textbf{81.6}
& \cellcolor{gg}\textbf{67.3}
& \cellcolor{gg}\textbf{75.8}
& \cellcolor{gg}\textbf{645}
& \cellcolor{gg}\textbf{40.5}
& \cellcolor{gg}\textbf{2170}
& \cellcolor{gg}\textbf{78.1}
& \cellcolor{gg}77.3
& \textbf{74.2}
& \textbf{88.9\%}\\
    \bottomrule
    \end{tabular}
  }
  \vspace{-5mm}
\end{table*}
\noindent\textbf{Layer-wise Sensitivity.}
At layer $l$, we estimate the local output sensitivity
to each visual token's hidden representation as
\begin{equation}
S_{\mathrm{layer}}^l(v_i)=
\left\|
\nabla_{H_i^l}F_{l+1}
\odot H_i^l
\right\|_2.
\label{eq:ctis_layer}
\end{equation}
The local gradient is obtained through a one-layer lookahead from $H^l$ to $H^{l+1}$. This captures cross-token influence within the lookahead layer without backpropagating through the remaining transformer layers. The layer-wise score measures the local output sensitivity to a token's hidden representation. It serves as a first-order proxy for replacing the selected hidden state with zero, rather than the total effect of intervening on the original input token.

\noindent\textbf{Score Normalization.}
To balance the association-corrected and layer-wise scores,
we standardize each component over the current retained
token set $\mathcal{S}$:
\begin{equation}
\mathcal{N}_{\mathcal{S}}(s_i)=
\frac{s_i-\mu_{\mathcal{S}}(s)}
{\sigma_{\mathcal{S}}(s)+\epsilon},
\label{eq:score_normalization}
\end{equation}
where $\mu_{\mathcal{S}}$ and $\sigma_{\mathcal{S}}$
denote the mean and standard deviation of the scores
over $\mathcal{S}$, respectively.

\noindent\textbf{Progressive Pruning Schedule.}
At each pruning layer $l\in\mathcal L_p$, we perform a one-layer lookahead to compute local sensitivity scores before removing low-ranked visual tokens. The lookahead layer is then recomputed using the retained tokens, and inference proceeds through the reduced sequence. The first pruning stage additionally computes the early gradient score and visual--text association correction. This avoids a full-model reference forward--backward pass while preserving progressive pruning across layers.

Algorithm~\ref{alg:causalpruning} summarizes CAPruner in the following three stages: (1) a forward--backward pass to compute gradient scores, (2) a calibration-based cross-modal association correction, and (3) layer-wise progressive pruning using combined sensitivity scores. The $O(Nd)$ gradient-score operations and
$O(M+NB_T)$ association correction exclude
the cost of prefix backpropagation and local
lookahead computations. These additional costs
must be included when evaluating end-to-end
inference efficiency.

\section{Experiments}
\subsection{Experimental Settings}
\subsubsection{Implementation Details}
We evaluate CAPruner on LLaVA-1.5-7B/13B~\cite{liu2024improved},
LLaVA-NeXT-7B~\cite{liu2024llavanext},
Qwen2.5-VL-7B~\cite{Qwen2.5-VL},
and InternVL3-8B~\cite{zhu2025internvl3}.
Following CDPruner~\cite{CDPruner}, we consider multiple
visual-token retention budgets and report performance
relative to the unpruned baseline.
CAPruner uses $|\mathcal{L}_p|=3$ pruning layers
(\textit{i.e.,} layers 8, 16, and 24),
a calibration size of $n=128$, a mixing coefficient of
$\alpha=0.6$, and $B_V=B_T=8$ relative-position bins.
The effects of $\alpha$, $B_V$, $B_T$, and $\mathcal{L}_p$
are examined in Section~\ref{hyperp}.
At the first pruning layer, we compute gradient-based token
scores from the early response via one-layer lookahead,
while subsequent pruning layers use block-local gradients
for progressive token selection.
For calibration, we randomly sample 128 image--question
pairs from the VQAv2 training split with a fixed seed of 42,
without using ground-truth answers.
For each backbone, we compute gradient-based visual and
textual sensitivity scores on the same calibration set,
aggregate them into $B_V$ and $B_T$ relative-position
bins, respectively, and estimate the association
coefficients offline.
The resulting coefficients are cached and reused during
inference without additional model training.
All experiments are training-free and conducted with
PyTorch 2.1 on eight NVIDIA RTX 4090 GPUs, using
\texttt{lmms-eval}~\cite{zhang2025lmms}.
Model forward and backward computations use BF16,
while sensitivity score computation and association
coefficient estimation use FP32 for numerical stability.
Further implementation details are provided in the
\textcolor{red}{Supplementary Material}.
\subsubsection{Benchmarks}
We evaluate CAPruner on a comprehensive broad suite of multimodal benchmarks that cover diverse aspects of image understanding, visual question answering, multimodal reasoning, hallucination evaluation, and text-centric visual comprehension. Specifically, we use 14 benchmarks. These include 10 general VQA and multimodal reasoning benchmarks, namely VQAv2~\cite{goyal2017vqav2}, GQA~\cite{hudson2019gqa}, VizWiz~\cite{gurari2018vizwiz}, ScienceQA-IMG~\cite{lu2022sqa}, HallBench~\cite{guan2024hallbench}, POPE~\cite{li2023pope}, MME~\cite{fu2024mme}, MMBench~\cite{liu2025mmbench}, MMBench-CN~\cite{liu2025mmbench}, and MM-Vet~\cite{yu2023mmvet}. We also include 4 text-oriented benchmarks, namely TextVQA~\cite{singh2019textvqa}, ChartQA~\cite{masry2022chartqa}, AI2D~\cite{kembhavi2016ai2d}, and OCRBench~\cite{liu2024ocrbench}, to assess OCR-related and document-level reasoning abilities across various visual scenarios. 
For the eight-benchmark Qwen2.5-VL evaluation, Avg. denotes the mean benchmark score after scaling OCRBench by 1/10 and MME by 1/20, while Rel. denotes the mean of per-benchmark performance ratios relative to the unpruned backbone.

\subsubsection{Comparison Methods}
We compare CAPruner with representative visual token pruning methods from three paradigms.
The attention-based methods include FastV~\cite{FastV}, SparseVLM~\cite{zhang2024sparsevlm}, 
PyramidDrop (PDrop)~\cite{pyramiddrop}, TRIM~\cite{TRIM}, VisionZip~\cite{visionzip}, 
and FlashVLM~\cite{cai2025flashvlm}. The gradient-based methods include ZOO-Prune~\cite{ZOOPrune} 
and FitPrune~\cite{ye2025fit}. The diversity-based methods include PruMerge+~\cite{PruMerge+}, 
DART~\cite{DART}, DivPrune~\cite{divprune}, CDPruner~\cite{CDPruner}, 
IDPruner~\cite{tan2026idpruner}, and HoloV~\cite{zou2026don}.
All methods follow official settings and metrics with unified token 
budgets and protocols for fair comparison.
\begin{figure*}[!t]
    \centering
    \includegraphics[width=1.0\textwidth]{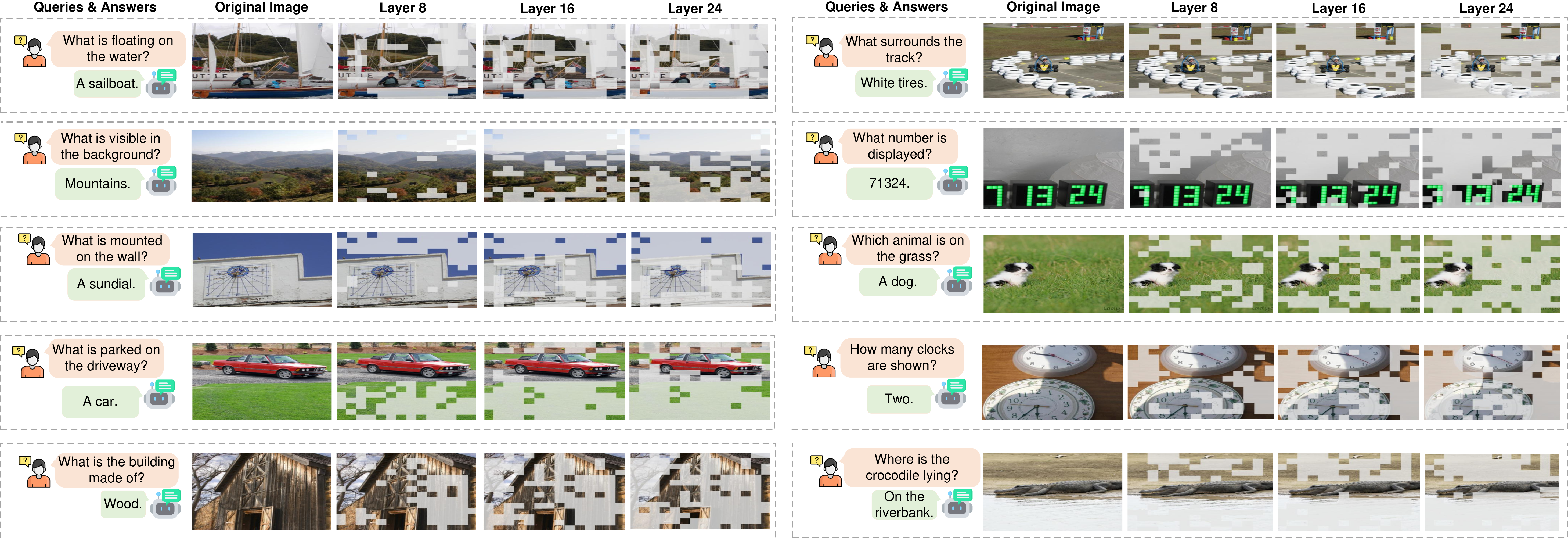}
\caption{Qualitative visualization of CAPruner on LLaVA-1.5-7B. Each case shows the original image and retained visual tokens after progressive pruning at Layers 8, 16, and 24.}
    \label{fig:viulize}
    \vspace{-2mm}
\end{figure*}
\begin{table*}[t]
\centering
\caption{Component ablations on Qwen2.5-VL-7B with 128 retained visual tokens. The best results are highlighted in \textbf{bold}}
\label{tab:components}
\resizebox{\textwidth}{!}{
\begin{tabular}{l|ccc|cccccccc|aa}
\toprule
\textbf{Configuration}
& \textbf{Assoc.}
& \textbf{Layer.}
& \textbf{Prog.}
& \textbf{TextVQA}
& \textbf{ChartQA}
& \textbf{AI2D}
& \textbf{OCRBench}
& \textbf{HallBench}
& \textbf{MME}
& \textbf{MMB-EN}
& \textbf{MMB-CN}
& \textbf{Avg.}
& \textbf{Rel.} \\

\rowcolor{ggr}
\multicolumn{14}{c}{
\textit{Keep 128 Tokens} \textcolor{Green}{(\textbf{Pruning Ratio} $\mathbf{90.1\%}$)}
} \\

Early GradScore (one-shot)
& -- & -- & --
& 80.2 & 63.5 & 74.2 & 617 & 38.1
& 2124 & 76.1 & 75.2 & 71.9 & 86.0\% \\

Early GradScore (progressive)
& -- & -- & $\checkmark$
& 80.4 & 64.3 & 74.8 & 622 & 38.7
& 2134 & 76.7 & 75.9 & 72.5 & 86.7\% \\

w/o Layer-wise Scoring
& $\checkmark$ & -- & $\checkmark$
& 80.8 & 65.8 & 75.0 & 634 & 39.2
& 2142 & 77.4 & 76.1 & 73.1 & 87.5\% \\

w/o Association Correction
& -- & $\checkmark$ & $\checkmark$
& 81.1 & 66.1 & 75.3 & 636 & \textbf{40.7}
& 2150 & 77.7 & 76.6 & 73.6 & 88.2\% \\
Full CAPruner (one-shot)
& $\checkmark$ & $\checkmark$ & --
& 80.8
& 66.2
& 74.8
& 632
& 39.0
& 2158
& 77.0
& \textbf{77.5}
& 73.3
& 87.7\% \\
\cellcolor{gg}\textbf{Full CAPruner} {\small(Ours)}
& \cellcolor{gg}$\checkmark$
& \cellcolor{gg}$\checkmark$
& \cellcolor{gg}$\checkmark$
& \cellcolor{gg}\textbf{81.6}
& \cellcolor{gg}\textbf{67.3}
& \cellcolor{gg}\textbf{75.8}
& \cellcolor{gg}\textbf{645}
& \cellcolor{gg}40.5
& \cellcolor{gg}\textbf{2170}
& \cellcolor{gg}\textbf{78.1}
& \cellcolor{gg}77.3
& \textbf{74.2}
& \textbf{88.9\%} \\

\bottomrule
\end{tabular}
}
\vspace{-1mm}
\end{table*}
\subsection{Main Results and Analysis}
Quantitative performance comparisons in the main paper focus on Qwen2.5-VL-7B. Additional evaluations on LLaVA-1.5-7B, LLaVA-1.5-13B, and InternVL3-8B, as well as video understanding benchmarks using Qwen2.5-VL-7B, are reported in the \textcolor{red}{Supplementary Material}. The end-to-end efficiency evaluation on LLaVA-NeXT-7B is presented in Section~\ref{efficiency}.

\subsubsection{Comparisons on Qwen2.5-VL-7B}
We evaluate CAPruner on Qwen2.5-VL-7B to assess its effectiveness on a dynamic-resolution MLLM. As shown in Table~\ref{tab:qwen2.5-vl}, CAPruner achieves the best average performance across token budgets. With 512 retained tokens, it preserves 98.2\% of the full-token performance and slightly outperforms CDPruner and FastV. When the budget is reduced to 256 tokens, CAPruner retains 93.9\% relative performance, exceeding CDPruner by 1.2 average-score points. Under the most aggressive 128-token setting, CAPruner maintains 88.9\% of the original performance and improves over CDPruner by 2.9 average points. The improvements on TextVQA, ChartQA, and MME further suggest that CAPruner can effectively preserve task-relevant visual information under aggressive token reduction.

\subsubsection{Qualitative Visualization}
We visualize the progressive pruning behavior of CAPruner on
LLaVA-1.5-7B using representative VQA examples, as shown in
Figure~\ref{fig:viulize}.
Each case presents the original image together with the visual
tokens retained after pruning at Layers 8, 16, and 24.
Across diverse query types, CAPruner progressively removes
less relevant visual regions while preserving compact evidence
relevant to the queried content.
For object- and attribute-centric questions, the retained
tokens remain concentrated on salient regions such as the
sailboat, sundial, car, dog, and wooden building.
For context-dependent queries, CAPruner additionally preserves
supporting visual evidence, such as the distant mountains,
the white tires surrounding the track, and the riverbank
beneath the crocodile.
For numerical recognition and counting tasks, the retained
tokens cover fine-grained visual details associated with
the digital display and the two clock faces, respectively.
These examples illustrate that CAPruner does not simply
retain foreground objects, but also preserves task-relevant
contextual and fine-grained visual evidence as pruning
progresses, resulting in interpretable token selections.

\subsection{Mechanism Analysis and Ablation Studies}

\subsubsection{Early Scoring Fidelity}
\label{early}
We examine whether early-stage sensitivity scores reflect the final model's response to visual-token interventions. On a held-out subset of 200 VQAv2 validation examples, we sample 32 visual tokens per example and compute exact early and final CTIS by individually replacing each selected token embedding with zero at its original position. Both responses use the same target $y^*$ selected by the unpruned Layer-9 readout, which is held fixed across interventions. We compare attention scores, ZOO-Prune's native sensitivity score, the absolute early directional derivative, and Early GradScore with different correction strategies, including static-bin, shuffled-text, multivariate ridge, and our association correction. Exact early CTIS and the full-output GradScore serve only as offline diagnostic references. The fidelity evaluation uses fixed-position interventions, while physical token deletion is examined separately below.
We report the mean per-example Spearman rank correlation with exact early and final CTIS, together with Recall@8 of the eight most influential tokens according to final CTIS among the same 32 sampled candidates.

We further measure the top-1 prediction agreement between
the Layer-9 readout and the final model and report
final-CTIS ranking correlations separately for matching
and mismatching predictions. To assess target mismatch,
we additionally evaluate ranking fidelity using the
final model's own unpruned top-1 prediction as the fixed
intervention target, without changing the online scores.
As shown in Table~\ref{tab:early_scoring_fidelity}, the association-corrected early score improves final-CTIS Spearman correlation from 0.59 to 0.67 and Recall@8 from 67.5\% to 73.9\% over the uncorrected GradScore. It also exceeds ZOO-Prune by 0.10 and 9.1
percentage points, respectively. Its advantages over static-bin and shuffled-text
corrections support the contribution of input-specific
textual sensitivities beyond fixed positional
calibration, while its competitive performance
against multivariate ridge correction supports
the adopted pairwise formulation. Its ranking fidelity approaches that of exact early CTIS (0.71), while the full-output gradient reference remains higher (0.76). 
The early and final predictions agree on 64.0\% of the evaluated examples, with final-CTIS Spearman correlations of 0.74 and 0.55 for matching and mismatching predictions, respectively. When using the final model's own prediction as the intervention target, association correction improves the Spearman correlation from 0.55 to 0.63 overall and from 0.44 to 0.51 on mismatching examples. These results suggest that the proposed scoring remains informative despite early-to-final prediction mismatch.
\begin{figure*}[!t]
    \centering
    \includegraphics[width=1\linewidth]{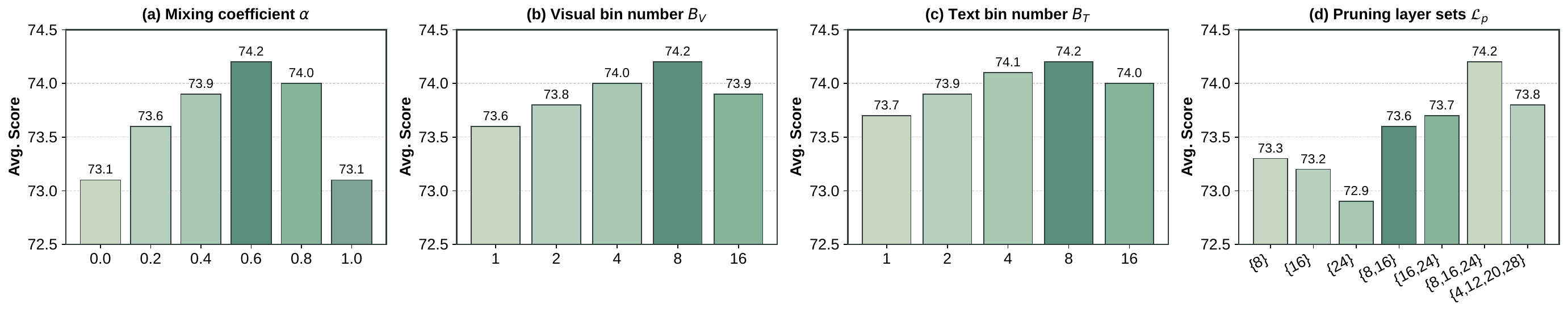}
\caption{
Hyperparameter analysis of CAPruner on Qwen2.5-VL-7B with 128 retained visual tokens.
(a) Effect of the mixing coefficient $\alpha$.
(b) Effect of the visual bin number $B_V$.
(c) Effect of the text bin number $B_T$.
(d) Effect of different pruning layer sets $\mathcal{L}_p$.
Results are reported in Avg. Score.
CAPruner achieves the best overall performance with $\alpha=0.6$, $B_V=8$, $B_T=8$, and $\mathcal{L}_p=\{8,16,24\}$.
}
  \label{hyper}
\end{figure*}
%
\begin{table}[!t]
\centering
\caption{Early scoring fidelity against exact token-intervention scores on LLaVA-1.5-7B. Recall@8 is evaluated over 32 sampled visual tokens.}
\label{tab:early_scoring_fidelity}
\resizebox{\columnwidth}{!}{
\begin{tabular}{l|ccc}
\toprule
\textbf{Scoring Method}
& \makecell[c]{\textbf{Early CTIS}\\\textbf{Spearman} $\uparrow$}
& \makecell[c]{\textbf{Final CTIS}\\\textbf{Spearman} $\uparrow$}
& \makecell[c]{\textbf{Final CTIS}\\\textbf{Recall@8} $\uparrow$} \\
\midrule
\rowcolor{ggr}
\multicolumn{4}{c}{\textit{Online scoring and calibration controls}} \\
Attention Score
& 0.29 & 0.25 & 40.4\% \\
ZOO-Prune (Sensitivity)
& 0.61 & 0.57 & 64.8\% \\
Early Directional Gradient
& 0.52 & 0.43 & 54.2\% \\
Early GradScore
& 0.68 & 0.59 & 67.5\% \\
+ Static Bin Correction
& 0.69 & 0.61 & 69.1\% \\
+ Shuffled Text Correction
& 0.67 & 0.58 & 66.4\% \\
+ Multivariate Ridge Correction
& 0.71 & 0.65 & 71.8\% \\
\rowcolor{gg}
\textbf{+ Association (Ours)}
& \textbf{0.72}
& \textbf{0.67}
& \textbf{73.9\%} \\
\midrule
\rowcolor{ggr}
\multicolumn{4}{c}{\textit{Offline diagnostic references}} \\
Exact Early CTIS
& 1.00 & 0.71 & 77.5\% \\
Full-output GradScore
& 0.59 & 0.76 & 81.8\% \\
\bottomrule
\end{tabular}
}
\vspace{-1mm}
\end{table}
\begin{table*}[t]
\centering
\caption{End-to-end inference efficiency on LLaVA-NeXT-7B with a maximum budget of 320 retained visual tokens. Latencies are in ms/sample. End-to-end timing includes visual processing and 32 fixed decoding steps.}
\label{tab:overhead}
\resizebox{\textwidth}{!}{%
\begin{tabular}{l|c|cccccc|a}
\toprule
\textbf{Method} & \textbf{Venue}
& \makecell{\textbf{Max. Vis.}\\\textbf{Tokens}}
& \makecell{\textbf{Total Prefill}\\\textbf{(ms)} $\downarrow$}
& \makecell{\textbf{Decode}\\\textbf{(ms/token)} $\downarrow$}
& \makecell{\textbf{End-to-End}\\\textbf{(ms)} $\downarrow$}
& \makecell{\textbf{Prefill Reduction}\\\textbf{(\%)} $\uparrow$}
& \makecell{\textbf{Peak GPU Memory}\\\textbf{(GB)} $\downarrow$}
& \makecell{\textbf{POPE F1}\\$\uparrow$} \\
\midrule
LLaVA-NeXT-7B & -- & 2880 & 246.0 & 29 & 1248.0 & -- & 16.7 & 86.8 \\
FastV & ECCV-2024 & 320 & 56.3 & 23 & 866.3 & 77.1\% & 15.6 & 49.5 \\
DivPrune & CVPR-2025 & 320 & \textbf{43.8} & 22 & 821.8 & \textbf{82.2\%} & \textbf{13.8} & 84.7 \\
CDPruner & NeurIPS-2025 & 320 & 45.4 & 22 & 823.4 & 81.5\% & \textbf{13.8} & 87.2 \\
\cellcolor{gg}\textbf{CAPruner}{\small\,(Ours)}
& \cellcolor{gg}This paper
& \cellcolor{gg}320
& \cellcolor{gg}130.0
& \cellcolor{gg}\textbf{19}
& \cellcolor{gg}\textbf{812.0}
& \cellcolor{gg}47.2\%
& \cellcolor{gg}16.5
& \textbf{88.8} \\
\bottomrule
\end{tabular}%
}
\vspace{-1mm}
\end{table*}
\begin{table}[t]
\centering
\caption{Breakdown of CAPruner's online prefill latency on LLaVA-NeXT-7B.}
\label{tab:CAPruner_overhead}
\resizebox{\columnwidth}{!}{%
\begin{tabular}{lca}
\toprule
\textbf{Component}
& \makecell{\textbf{Latency}\\\textbf{(ms/sample)}}
& \textbf{Share} \\
\midrule
Unpruned prefix forward (Layers 1--9)
& 40.0 & 30.77\% \\
Prefix backward (Layers 1--9)
& 36.0 & 27.69\% \\
Later one-layer lookahead forward
& 8.0 & 6.15\% \\
Local backward (Layers 17 and 25)
& 12.0 & 9.23\% \\
Lookahead replay (Layers 9, 17, and 25)
& 9.0 & 6.92\% \\
Association correction and Top-$k$
& 2.0 & 1.54\% \\
Remaining pruned forward
& 23.0 & 17.69\% \\
\midrule
\rowcolor{gg}\textbf{Total online LLM prefill}
& \textbf{130.0}
& \textbf{100.00\%} \\
\bottomrule
\end{tabular}%
}
\vspace{-5mm}
\end{table}

\subsubsection{Masking vs. Token Deletion}
To examine whether fixed-position token masking reflects actual token removal, we additionally compare exact zero-masking CTIS with the output changes induced by individually deleting the same visual tokens. Both interventions are applied at the same visual-token interface, with positional metadata and attention masks handled consistently with the corresponding intervention. Using the same sampled candidates and fixed response target, we measure the per-example rank correlation between masking- and deletion-based sensitivities, as well as the correlations between gradient-based scores and deletion-based sensitivities. The resulting mean Spearman correlation between masking- and deletion-based sensitivities is 0.82. Our association-corrected score achieves a correlation of 0.63 with deletion-based sensitivity, compared with 0.55 for the uncorrected Early GradScore. These results suggest that fixed-position masking provides a useful reference for token-deletion sensitivity, while association correction improves the alignment of online scores with actual token-removal effects.

\subsubsection{Component Ablations}
We evaluate the contribution of each CAPruner component on Qwen2.5-VL-7B with 128 retained visual tokens (90.1\% pruning), using the same evaluation protocol as Table~\ref{tab:qwen2.5-vl}. We independently examine cross-modal association correction, layer-wise sensitivity scoring, and progressive pruning while keeping the final token budget and other settings fixed. 
The one-shot variants prune directly to the target budget at the first pruning layer, while the progressive variants distribute token removal across multiple layers. We additionally compare the full configuration with its one-shot counterpart, which retains association correction and first-stage layer-wise scoring but disables subsequent pruning and score updates. 
Without layer-wise scoring, the initial association-corrected ranking is reused at subsequent pruning stages.
As shown in Table~\ref{tab:components}, progressive pruning improves the average score from 71.9 to 72.5 under the same early-gradient ranking. Removing layer-wise sensitivity or association correction from the full configuration results in average scores of 73.1 and 73.6, respectively, compared with 74.2 for full CAPruner. 
These results highlight the complementary contributions of cross-modal correction and depth-dependent sensitivity estimation to visual token selection.
Compared with its one-shot counterpart, full CAPruner improves the average score from 73.3 to 74.2, suggesting the benefit of progressive pruning with depth-dependent sensitivity updates.
\subsubsection{Hyperparameter Analysis}
\label{hyperp}
We analyze the sensitivity of CAPruner to four key hyperparameters on Qwen2.5-VL-7B with 128 retained visual tokens, including the mixing coefficient $\alpha$, the visual and text bin numbers $B_V$ and $B_T$, and the pruning layer set $\mathcal{L}_p$. As shown in Fig.~\ref{hyper}, CAPruner is reasonably stable across a broad range of configurations, while moderate settings consistently provide the best performance. In particular, the best result is achieved with $\alpha=0.6$, indicating that the association-corrected early score and the layer-wise sensitivity score are complementary. For the association correction, both overly coarse and overly fine binning are suboptimal, and the best performance is obtained at $B_V=8$ and $B_T=8$. For progressive pruning, multi-stage pruning is clearly more effective than single-stage pruning, while the three-stage configuration $\mathcal{L}_p=\{8,16,24\}$ performs best overall.

\subsection{End-to-End Inference Efficiency}
\label{efficiency}
\subsubsection{End-to-End Latency}
We evaluate the inference efficiency of CAPruner on 
LLaVA-NeXT-7B using the POPE benchmark, with a maximum 
budget of 320 retained visual tokens, and compare it 
against the full-token backbone and representative 
pruning baselines. All methods are profiled with batch 
size one under the same input, precision, hardware, 
and decoding settings. We measure total LLM prefill 
latency, including online token scoring and pruning, 
as well as model-side end-to-end latency encompassing 
visual processing, prefill, and 32 fixed decoding steps. 
Peak GPU memory is measured over the complete online 
inference pipeline, including gradient-based scoring 
and decoding, while POPE F1 is also reported.
As illustrated in Table~\ref{tab:overhead}, CAPruner 
reduces total prefill latency from 246.0 to 130.0 ms 
(47.2\%) and model-side end-to-end latency from 
1248.0 to 812.0 ms (34.9\%) relative to the unpruned 
backbone. Compared with CDPruner, CAPruner achieves 
a higher POPE F1 (88.8 \textit{vs.} 87.2) while 
maintaining slightly lower end-to-end latency 
(812.0 \textit{vs.} 823.4 ms). Although gradient-based 
scoring introduces additional prefill and memory 
overhead compared with lightweight pruning baselines, 
CAPruner achieves the lowest end-to-end latency and 
the highest POPE F1 among the evaluated methods, 
demonstrating a favorable accuracy--efficiency trade-off.

\subsubsection{Computational Overhead Breakdown}
To identify the online cost of the scoring pipeline,
we profile the unpruned prefix forward and backward, later
one-layer lookahead computations, local backward passes,
replay after pruning, association correction and token
selection, and the remaining pruned forward computation.
The initial Layer-9 lookahead is included in the prefix
forward, while the later lookahead forward includes only
Layers 17 and 25. Thus, no computation is counted twice.
As shown in Table~\ref{tab:CAPruner_overhead},
the unpruned prefix forward accounts for the largest
individual cost (40.0 ms, 30.77\% of prefill), while
the combined backward operations contribute 48.0 ms
(36.92\%). In contrast, association correction and
token selection incur only 2.0 ms (1.54\%).
The total equals the 130.0 ms prefill
time reported in Table~\ref{tab:overhead}.
Offline estimation of the association coefficients
is excluded from online latency.
These results indicate that gradient computation remains
a major source of additional overhead, whereas association
correction and token selection introduce relatively
little computational cost.
\section{Conclusion}
In this paper, we present CAPruner, an intervention-inspired visual token pruning framework for efficient MLLMs. CAPruner uses fixed-input token perturbations as a reference for evaluating response sensitivity and develops an early-readout gradient-based scoring strategy, refined by calibrated cross-modal association correction and complemented by one-layer-lookahead hidden-state sensitivity estimates for progressive pruning.
These scores serve as computational proxies for model-internal intervention sensitivity rather than identified population-level causal effects. Experiments across multiple MLLM architectures and vision-language benchmarks show favorable performance retention under aggressive visual token compression, while analyses of both scoring fidelity and end-to-end inference efficiency further assess the reliability and computational cost of CAPruner.

\printbibliography




\clearpage

\makeatletter
\let\maketitle\savedmaketitle
\let\@maketitle\saved@maketitle
\let\thanks\savedthanks
\makeatother

\title{Supplementary Materials for Beyond Correlation: Causality-Driven Visual Token Reduction for MLLMs via Counterfactual Intervention}

\author{Pengcheng Zheng, Chaoning Zhang, \IEEEmembership{Senior Member, IEEE}, Ya Wen, Jiaxin Huang, Jiaquan Zhang, Caiyan Qin, Kuien Liu, Sheng Zheng, Pietro Lió, \IEEEmembership{Fellow, IEEE}, and Yang Yang, \IEEEmembership{Senior Member, IEEE}
\thanks{This work was supported by the National Natural Science Foundation of China under grant 62572104 and 62220106008.} 
\thanks{Pengcheng Zheng, Chaoning Zhang, Ya Wen, Jiaquan Zhang, Sheng Zheng, and Yang Yang are with the Center for Future Media and School
of Computer Science and Engineering, University of Electronic Science and Technology of China, Chengdu 611731, China (e-mail: zpc777@std.uestc.edu.cn; chaoningzhang1990@gmail.com).}
\thanks{Jiaxin Huang is with the Department of Machine Learning, Mohamed bin Zayed University of Artificial Intelligence, Abu Dhabi, United Arab Emirates.}
\thanks{Caiyan Qin is with the School of Robotics and Advanced Manufacture, Harbin Institute of Technology, Shenzhen 518055, China.}
\thanks{Kuien Liu is with the Institute of Software, Chinese Academy of Sciences, Beijing 100190, China}
\thanks{Pietro Lió is with the Department of Computer Science and Technology, University of Cambridge, GB-CAM Cambridge, U.K.}
}

\markboth{Journal of \LaTeX\ Class Files,~Vol.~14, No.~8, August~2021}%
{Shell \MakeLowercase{\textit{et al.}}: A Sample Article Using IEEEtran.cls for IEEE Journals}

\maketitle

\begin{abstract}
Here, we provide supplementary materials to further support and validate the proposed CAPruner framework. The supplement includes the MLLM architecture and token
reduction setup, a detailed proof of the early gradient-based
sensitivity bound, additional experimental setup details,
extended evaluations on LLaVA-1.5-7B, LLaVA-1.5-13B, and InternVL3-8B, and a generalization evaluation on video understanding benchmarks using Qwen2.5-VL-7B.
\end{abstract}

\begin{IEEEkeywords}
Multimodal large language models, counterfactual intervention, visual token reduction, efficient inference.
\end{IEEEkeywords}

\section{MLLM Architecture and Token Reduction Setup}
An MLLM consists of three components: a visual encoder $\mathcal{V}$ (typically a CLIP
ViT~\cite{radford2021learning}), a projection layer $\mathcal{P}$, and an LLM backbone $\mathcal{L}$. Given an image $I$ and a text query $q$, the visual encoder
produces $N$ visual token embeddings $\mathbf{V} = \{v_1, \ldots, v_N\} \in \mathbb{R}^{N \times d}$,
which are projected by $\mathcal{P}$ and concatenated with text token embeddings
$\mathbf{T} = \{t_1, \ldots, t_M\} \in \mathbb{R}^{M \times d}$ to form the input sequence
$\mathbf{X} = [\mathbf{V}; \mathbf{T}] \in \mathbb{R}^{(N+M) \times d}$ fed into $\mathcal{L}$.
The LLM backbone processes $\mathbf{X}$ through $L$ transformer layers, each consisting
of multi-head self-attention and feed-forward sublayers~\cite{vaswani2017attention}, producing output logits
$Y = \mathcal{L}(\mathbf{X}) \in \mathbb{R}^{|\mathcal{V}_{\text{vocab}}|}$.
The token reduction problem is defined as follows. Given a token retention ratio $\rho \in (0,1)$, select a subset $\mathcal{S} \subseteq [N]$ with
$|\mathcal{S}| = \lfloor \rho N \rfloor$ such that the model output $Y_\mathcal{S}$
computed with only the retained tokens is maximally close to the full-token output
$Y_{[N]}$. Formally, we seek:
\begin{equation}
\mathcal{S}^* = \arg\min_{\mathcal{S} \subseteq [N],\, |\mathcal{S}|=\lfloor\rho N\rfloor}
\bigl\| Y_\mathcal{S} - Y_{[N]} \bigr\|_2
\end{equation}
This is a combinatorial optimization problem that is NP-hard in general, so practical
methods rely on importance scores to greedily select the top-$k$ tokens.


\begin{table*}[!t]
  \centering
\caption{
Main results on LLaVA-1.5-7B under different visual token budgets. \textbf{Acc.} denotes the average performance across diverse benchmarks, and \textbf{Rel.} denotes
the mean of per-benchmark performance ratios relative to
the full-token baseline. The best results are highlighted in \textbf{bold}. Aggregate metrics are omitted for methods with
incomplete benchmark results.
}
  \label{tab:llava-1.5}
  \resizebox{\linewidth}{!}{
    \begin{tabular}{l|c|cccccccccc|aa}
    \toprule
    \textbf{Method} & \textbf{Venue} &\textbf{$\text{VQA}^\text{V2}$} & \textbf{GQA} & \textbf{VizWiz} & \textbf{$\text{SQA}^\text{IMG}$} & \textbf{$\text{VQA}^\text{Text}$} & \textbf{POPE} & \textbf{MME} & \textbf{$\text{MMB}^\text{EN}$} & \textbf{$\text{MMB}^\text{CN}$} & \textbf{MMVet} & \textbf{Acc.} & \textbf{Rel.} \\
    \rowcolor{ggr}\multicolumn{14}{c}{\textit{Full-token Baseline, All 576 Tokens} ($\mathbf{100\%}$)} \\
     LLaVA-1.5-7B &- & 78.5 & 61.9 & 50.1 & 69.5 & 58.2 & 85.9 & 1506.5 & 64.7 & 58.1 & 31.3 & 63.4 & 100.0\% \\
    \rowcolor{ggr}\multicolumn{14}{c}{\textit{Keep 128 Tokens} \textcolor{Green}{(\textbf{Pruning Ratio} $\mathbf{77.8\%}$)}} \\
     FastV &ECCV-2024 & 71.0 & 54.0 & 51.9 & 69.2 & 56.4 & 68.2 & 1368.9 & 63.0 & 55.9 & 27.0 & 58.5 &  92.8\% \\
     FlashVLM &arXiv-2025 & 76.4 & 58.9 & 53.1 & 69.5 & 57.6 & 85.1 & 1483.2 & 63.2 & \textbf{57.6} & \textbf{34.1} & 63.0 &  100.1\% \\
     IDPruner &arXiv-2026 & 73.5 & 56.0 & 51.0 & 66.9 & 46.4 & 84.7 & 1428.0 & 60.4 & 49.2 & 28.0 & 58.8 &  92.3\% \\
     HoloV &NeurIPS-2025 & 75.5 & 57.7 & 51.5 & 69.8 & 56.8  & 84.0  & \textbf{1802.0}  & \textbf{63.9} & 56.5 & 24.5 & 63.0 &  98.2\% \\
     PDrop & CVPR-2025 & 74.3 & 57.1 & 49.4 & \textbf{70.1} & 56.7 & 77.5 & 1444.1 & 62.3 & 55.3 & 27.6 & 60.3 &  95.0\% \\
     SparseVLM & ICML-2025 & 75.1 & 57.3 & 49.7 & 69.0 & 56.3 & 83.1 & 1399.3 & 62.6 & 56.9 & 29.7 & 61.0 &  96.3\% \\
     PruMerge+ &ICCV-2025 & 75.0 & 58.2 & 53.7 & 69.1 & 54.0 & 83.1 & 1408.1 & 61.8 & 55.8 & 30.4 & 61.2 &  96.8\% \\
     TRIM & COLING-2025 & 75.4 & 58.4 & 51.6 & 68.6 & 52.2 & 85.3 & 1413.4 & 63.0 & 52.3 & 29.9 & 60.7 &  95.8\% \\
     VisionZip &CVPR-2025 & 75.6 & 57.6 & 51.6 & 68.7 & 56.9 & 83.3 & 1436.9 & 62.1 & 57.0 & 31.6 & 61.6 &  97.6\% \\
     DART &EMNLP-2025 & 74.7 & 57.9 & 52.8 & 69.1 & 56.3 & 80.4 & 1408.7 & 60.7 & 57.3 & 30.9 & 61.1 &  96.9\% \\
     DivPrune &CVPR-2025 & 76.0 & 59.4 & 52.8 & 68.6 & 55.9 & 87.0 & 1405.1 & 61.5 & 54.8 & 30.6 & 61.7 &  97.5\% \\
     ZOO-Prune &CVPR-2026 & 76.6 & 59.5 & - & 68.9 & 57.9 & 87.1 & - & 61.9 & - & - & -- & -- \\
     CDPruner &NeurIPS-2025 &76.6 & 59.9 & 52.8 & 69.0 & 56.2 & 87.7 & 1431.4 & 63.1 & 55.0 & 32.8 & 62.5 &  99.0\% \\\cellcolor{gg}\textbf{CAPruner}{ \small(Ours)} & \cellcolor{gg}This paper&\cellcolor{gg}\textbf{77.0} & \cellcolor{gg}\textbf{60.0} & \cellcolor{gg}\textbf{54.1} & \cellcolor{gg}69.5 & \cellcolor{gg}\textbf{58.2} & \cellcolor{gg}\textbf{88.5} & \cellcolor{gg}1446.7 & \cellcolor{gg}63.0 & \cellcolor{gg}56.9 & \cellcolor{gg}33.4 & \textbf{63.3} &\textbf{100.4\%} \\
    \rowcolor{ggr}\multicolumn{14}{c}{\textit{Keep 64 Tokens} \textcolor{Green}{(\textbf{Pruning Ratio} $\mathbf{88.9\%}$)}} \\
     FastV &ECCV-2024 & 55.9 & 46.0 & 49.1 & \textbf{70.1} & 51.6 & 35.5 &  973.5 & 50.1 & 42.1 & 18.9 & 46.8 &  74.9\% \\
     FlashVLM &arXiv-2025 & 73.6 & 56.1 & 53.6 & 69.3 & \textbf{56.1} & 81.7 & 1383.4 & 61.8 & \textbf{55.8} & 32.9 & 61.0 &  97.1\% \\
    IDPruner &arXiv-2026 & 71.0 & 53.5 & 50.5 & 66.7 & 44.7 & 83.8 & 1329.0 & 57.7 & 47.0 & 23.7 & 56.5 &  88.2\% \\
    HoloV &NeurIPS-2025 & 72.8 & 55.3 & 52.8 & 64.5 & 50.4
    & 78.3 & \textbf{1715.0} & \textbf{63.3} & 55.1
    & \textbf{36.8} & 61.5 & 98.2\% \\
     PDrop &CVPR-2025 & 56.3 & 46.1 & 46.3 & 68.8 & 49.2 & 40.8 &  982.2 & 48.0 & 36.6 & 17.7 & 45.9 &  72.9\% \\
     SparseVLM &ICML-2025 & 66.9 & 52.0 & 49.4 & 69.2 & 52.1 & 69.7 & 1190.4 & 58.3 & 49.6 & 24.4 & 55.1 &  87.1\% \\
     PruMerge+ &ICCV-2025 & 71.3 & 55.4 & \textbf{53.7} & 69.5 & 52.0 & 75.7 & 1316.8 & 59.6 & 52.1 & 28.0 & 58.3 &  92.4\% \\
     TRIM &COLING-2025 & 72.4 & 56.6 & 51.1 & 69.0 & 49.7 & 85.9 & 1350.9 & 60.9 & 48.2 & 24.8 & 58.6 &  91.6\% \\
     VisionZip &CVPR-2025 & 72.4 & 55.1 & 52.9 & 69.0 & 55.5 & 77.0 & 1365.2 & 60.1 & 55.4 & 29.4 & 59.5 &  94.4\% \\
     DART &EMNLP-2025 & 71.3 & 54.7 & 53.5 & 69.3 & 54.7 & 73.8 & 1365.1 & 59.5 & 54.0 & 26.5 & 58.6 &  92.6\% \\
     DivPrune &CVPR-2025 & 74.1 & 57.5 & 53.6 & 68.0 & 54.5 & 85.5 & 1334.7 & 60.1 & 52.3 & 28.1 & 60.0 &  94.7\% \\
     ZOO-Prune &CVPR-2026 & 75.0 & 58.5 & - & 68.3 & 55.3 & 85.9 & - & 60.2 & - & - & -- & -- \\
     CDPruner &NeurIPS-2025 & 75.4 & 58.6 & 53.4 & 68.1 & 55.3 & 87.5 & 1415.1 & 61.1 & 53.2 & 30.5 & 61.4 &  97.0\% \\
     \cellcolor{gg}\textbf{CAPruner}{ \small(Ours)} & \cellcolor{gg}This paper&\cellcolor{gg}\textbf{76.0} & \cellcolor{gg}\textbf{58.8} & \cellcolor{gg}53.6 & \cellcolor{gg}68.8 & \cellcolor{gg}55.6 & \cellcolor{gg}\textbf{88.3} & \cellcolor{gg}1495.7 & \cellcolor{gg}61.8 & \cellcolor{gg}54.5 & \cellcolor{gg}31.2 & \textbf{62.3} & \textbf{98.4\%} \\
    \rowcolor{ggr}\multicolumn{14}{c}{\textit{Keep 32 Tokens} \textcolor{Green}{(\textbf{Pruning Ratio} $\mathbf{94.4\%}$)}} \\
    FlashVLM &arXiv-2025 & 69.3  & 52.8 & 53.2 & 69.4 & 54.6 & 74.5 & 1286.5 & 59.3 & \textbf{53.4} & \textbf{29.6} & 58.0 &  92.4\% \\
    IDPruner &arXiv-2026 & 68.5 & 51.0 & 49.5 & 66.7 & 41.5 & 81.4 & 1254.0 & 55.5 & 44.8 & 20.5 & 54.2 &  84.2\% \\
     PruMerge+ & ICCV-2025 & 65.6 & 52.9 & \textbf{53.5} & 67.9 & 49.2 & 66.7 & 1236.6 & 55.1 & 45.9 & 24.7 & 54.3 & 86.1\% \\
    TRIM &COLING-2025 & 68.6 & 54.5 & 50.7 & 68.1 & 47.6 & 84.9 & 1251.8 & 57.7 & 40.1 & 20.5 & 55.5 & 86.2\% \\
    VisionZip &CVPR-2025 & 67.1 & 51.8 & 52.4 & 69.1 & 53.1 & 69.4 & 1251.2 & 57.0 & 50.3 & 25.3 & 55.8 & 88.4\% \\
    DART &EMNLP-2025 & 67.1 & 52.9 & 52.5 & 69.3 & 52.2 & 69.1 & 1273.3 & 58.5 & 50.0 & 25.0 & 56.0 & 88.6\% \\
    DivPrune &CVPR-2025 & 71.2 & 54.9 & 53.3 & 68.6 & 52.9 & 81.5 & 1284.9 & 57.6 & 49.1 & 26.3 & 58.0 & 91.3\% \\
    CDPruner &NeurIPS-2025  &73.6 & 57.0 & 53.1 & 69.5 & 53.2 & 87.9 & 1373.0 & 59.6 & 49.6 & 27.8 & 60.0 &  94.3\% \\
    \cellcolor{gg}\textbf{CAPruner}{ \small(Ours)} &\cellcolor{gg}This paper &\cellcolor{gg}\textbf{74.9}  & \cellcolor{gg}\textbf{59.1} & \cellcolor{gg}\textbf{53.5} & \cellcolor{gg}\textbf{70.0} & \cellcolor{gg}\textbf{54.8} & \cellcolor{gg}\textbf{88.0} & \cellcolor{gg}\textbf{1439.4} & \cellcolor{gg}\textbf{60.4} & \cellcolor{gg}50.3 & \cellcolor{gg}28.4 & \textbf{61.1} & \textbf{96.1\%} \\
    \bottomrule
    \end{tabular}
  }
  \vspace{-3mm}
\end{table*}
\section{Proof of Theorem 2}

\begin{proof}
Let $V=\{v_1,\ldots,v_N\}$ denote the visual-token embeddings, with $v_i\in\mathbb{R}^d$, and let $F_e(V,T)=\log p_{\ell_0+1}(y^*\mid V,T)$ be the early scalar response defined in the main paper, where $y^*$ is selected from the unpruned early readout and remains fixed throughout the intervention. For a given token $v_i$, consider the intervention path $V_i(t)=V^{(i\leftarrow tv_i)}$, $t\in[0,1]$, which satisfies $V_i(1)=V$ and $V_i(0)=V^{(i\leftarrow0)}$, while all other visual and textual tokens remain unchanged. Defining $g_i(t)=F_e(V_i(t),T)$, the first- and second-order derivatives along this path are
\begin{equation}
\begin{aligned}
g_i'(t)
&=\nabla_{v_i}F_e(V_i(t),T)^\top v_i,\\
g_i''(t)
&=v_i^\top\nabla_{v_i}^2F_e(V_i(t),T)v_i.
\end{aligned}
\end{equation}
Assuming that $F_e$ is twice continuously differentiable along the entire intervention path considered here, Taylor's theorem~\cite{nocedal2006numerical} with integral remainder gives
\begin{equation}
\begin{aligned}
F_e(V,T)-F_e(V^{(i\leftarrow0)},T)
&=\nabla_{v_i}F_e(V,T)^\top v_i+r_i,\\
r_i
&=-\int_0^1 t\,g_i''(t)\,dt.
\end{aligned}
\end{equation}
Under the assumption that the Hessian spectral norm satisfies $\|\nabla_{v_i}^2F_e(V_i(t),T)\|_2\leq\Lambda_i$ for all $t\in[0,1]$, the remainder can be bounded as
\begin{equation}
\begin{aligned}
|r_i|
&\leq\int_0^1 t\,|g_i''(t)|\,dt\\
&\leq\int_0^1 t\,\Lambda_i\|v_i\|_2^2\,dt\\
&=\frac{\Lambda_i}{2}\|v_i\|_2^2.
\end{aligned}
\end{equation}
It follows that
\begin{equation}
\begin{aligned}
\mathrm{CTIS}_e(v_i)
&=\left|F_e(V,T)-F_e(V^{(i\leftarrow0)},T)\right|\\
&\leq\left|\nabla_{v_i}F_e(V,T)^\top v_i\right|
+\frac{\Lambda_i}{2}\|v_i\|_2^2.
\end{aligned}
\end{equation}
Let $a_i=\nabla_{v_i}F_e(V,T)\odot v_i\in\mathbb{R}^d$. By the Cauchy--Schwarz inequality~\cite{fikhtengol2014fundamentals}, $|\nabla_{v_i}F_e(V,T)^\top v_i|=|\sum_{k=1}^{d}(a_i)_k|\leq\sqrt{d}\|a_i\|_2=\sqrt{d}\,\mathrm{GradScore}(v_i)$. Combining this inequality with the remainder bound yields
\begin{equation}
\mathrm{CTIS}_e(v_i)
\leq
\sqrt{d}\,\mathrm{GradScore}(v_i)
+\frac{\Lambda_i}{2}\|v_i\|_2^2.
\end{equation}
The bound characterizes the relationship between the early gradient-based sensitivity score and the corresponding early intervention response. It does not imply equality between these quantities or guarantee approximation accuracy or ranking consistency with the final-model CTIS.
\end{proof}
\begin{table*}[t]
  \centering
  \caption{
Main results on LLaVA-1.5-13B under different visual token budgets. \textbf{Acc.} denotes the average performance across diverse benchmarks, and \textbf{Rel.} denotes
the mean of per-benchmark performance ratios relative to
the full-token baseline. The best results are highlighted in \textbf{bold}. Aggregate metrics are omitted for methods with
incomplete benchmark results.
}
  \label{tab:LLaVA-1.5-13B}
  \resizebox{\linewidth}{!}{
    \begin{tabular}{l|c|cccccccccc|aa}
    \toprule
    \textbf{Method} &Venue & \textbf{$\text{VQA}^\text{V2}$} & \textbf{GQA} & \textbf{VizWiz} & \textbf{$\text{SQA}^\text{IMG}$} & \textbf{$\text{VQA}^\text{Text}$} & \textbf{POPE} & \textbf{MME} & \textbf{$\text{MMB}^\text{EN}$} & \textbf{$\text{MMB}^\text{CN}$} & \textbf{MMVet} & \textbf{Acc.} & \textbf{Rel.} \\
    \rowcolor{ggr}\multicolumn{14}{c}{\textit{Full-token Baseline, All 576 Tokens} ($\mathbf{100\%}$)} \\
    LLaVA-1.5-13B &- & 80.0 & 63.3 & 53.6 & 72.8 & 61.2 & 86.0 & 1531.2 & 68.5 & 63.5 & 36.2 & 66.2 & 100.0\% \\
    \rowcolor{ggr}\multicolumn{14}{c}{\textit{Keep 128 Tokens} \textcolor{Green}{(\textbf{Pruning Ratio} $\mathbf{77.8\%}$)}} \\
    FastV &ECCV-2024 & 75.3 & 58.3 & \textbf{54.6} & 74.2 & 58.6 & 75.5 & 1460.6 & 66.1 & 62.3 & 32.8 & 63.1 &  95.4\% \\
    PDrop &CVPR-2025 & 78.2 & \textbf{61.0} & 53.8 & 73.3 & \textbf{60.2} & 83.6 & 1489.5 & 67.5 & \textbf{62.8} & 32.1 & 64.7 &  97.4\% \\
    SparseVLM &ICML-2025 & 77.6 & 59.6 & 51.4 & \textbf{74.3} & 59.3 & 85.0 & 1487.9 & \textbf{68.4} & 62.6 & 35.2 & 64.8 &  97.8\% \\
    PruMerge+ &ICCV-2025 & 76.2 & 58.3 & 52.8 & 73.3 & 56.1 & 82.7 & 1445.9 & 66.3 & 61.2 & 33.6 & 63.3 &  95.5\% \\
    TRIM & COLING-2025 & 76.4 & 59.4 & 49.7 & 72.4 & 55.0 & 86.8 & 1426.9 & 67.1 & 58.4 & 35.1 & 63.2 &  95.2\% \\
    VisionZip & CVPR-2025 & 76.8 & 57.9 & 52.3 & 73.8 & 58.9 & 82.7 & 1449.2 & 67.4 & 62.5 & 36.0 & 64.1 &  97.0\% \\
    DART &EMNLP-2025 & 75.7 & 57.7 & 53.0 & 74.2 & 58.7 & 80.4 & 1395.0 & 65.4 & 62.2 & 34.8 & 63.2 &  95.7\% \\
    DivPrune & CVPR-2025 & 77.1 & 59.2 & 53.5 & 72.8 & 58.0 & 86.8 & 1457.7 & 66.3 & 60.7 & 34.4 & 64.2 &  96.8\% \\
    ZOO-Prune & CVPR-2026 & 77.8 & 58.9 & - & 73.4 & 58.8 & 87.0 & - & 67.0 & - & - & -- & -- \\
    CDPruner &NeurIPS-2025 & 77.7 & 59.7 & 52.9 & 73.2 & 58.4 & 87.3 & 1478.0 & 67.5 & 61.5 & 36.2 & 64.8 & 98.0\% \\
\cellcolor{gg}\textbf{CAPruner}{ \small(Ours)} & \cellcolor{gg}This paper
& \cellcolor{gg}\textbf{78.6} 
& \cellcolor{gg}60.7 
& \cellcolor{gg}53.5 
& \cellcolor{gg}73.7 
& \cellcolor{gg}59.3 
& \cellcolor{gg}\textbf{88.0} 
& \cellcolor{gg}\textbf{1492.6} 
& \cellcolor{gg}67.4 
& \cellcolor{gg}62.2 
& \cellcolor{gg}\textbf{37.0} 
& \textbf{65.5} 
& \textbf{99.0\%} \\
    \rowcolor{ggr}\multicolumn{14}{c}{\textit{Keep 64 Tokens} \textcolor{Green}{(\textbf{Pruning Ratio} $\mathbf{88.9\%}$)}} \\
    FastV &ECCV-2024 & 65.3 & 51.9 & 53.8 & 73.1 & 53.4 & 56.9 & 1246.4 & 59.2 & 55.1 & 26.9 & 55.8 &  84.7\% \\
    PDrop &CVPR-2025 & 70.8 & 54.1 & 50.5 & 73.1 & 55.3 & 66.1 & 1247.0 & 63.1 & 56.6 & 21.9 & 57.4 &  85.9\% \\
    SparseVLM &ICML-2025 & 73.2 & 55.9 & 52.1 & 73.0 & 57.1 & 77.9 & 1374.3 & 65.2 & 60.3 & 32.9 & 61.6 &  93.2\% \\
    PruMerge+ &ICCV-2025 & 72.6 & 56.3 & 52.4 & 73.5 & 54.4 & 75.7 & 1338.2 & 65.0 & 59.3 & 30.3 & 60.6 &  91.5\% \\
    TRIM &COLING-2025 & 73.2 & 57.9 & 49.2 & 72.0 & 52.0 & 86.5 & 1406.2 & 65.0 & 52.7 & 27.8 & 60.7 &  90.6\% \\
    VisionZip &CVPR-2025 & 73.7 & 56.2 & 53.2 & \textbf{74.2} & 57.4 & 75.7 & 1379.6 & 64.9 & \textbf{61.3} & 33.4 & 61.9 &  93.8\% \\
    DART & EMNLP-2025 & 72.4 & 55.7 & 53.4 & 73.8 & 57.3 & 72.8 & 1380.0 & 64.7 & 60.6 & 32.8 & 61.3 &  92.8\% \\
    DivPrune &CVPR-2025 & 75.2 & 57.9 & \textbf{54.4} & 71.7 & 57.4 & 84.5 & 1454.2 & 64.1 & 59.8 & 29.3 & 62.7 &  94.1\% \\
    ZOO-Prune & CVPR-2026 & 76.4 & 58.6 & - & 72.1 & 58.6 & 85.3 & - & 64.8 & - & - & -- & -- \\
    CDPruner &NeurIPS-2025 & 76.7 & 59.4 & 53.6 & 72.5 & 57.6 & 87.1 & 1466.8 & 65.5 & 58.8 & 35.2 & 64.0 & 96.6\% \\
\cellcolor{gg}\textbf{CAPruner}{ \small(Ours)} & \cellcolor{gg}This paper
& \cellcolor{gg}\textbf{77.7} 
& \cellcolor{gg}\textbf{60.2} 
& \cellcolor{gg}\textbf{54.4} 
& \cellcolor{gg}73.3 
& \cellcolor{gg}\textbf{58.7} 
& \cellcolor{gg}\textbf{88.2} 
& \cellcolor{gg}\textbf{1484.0} 
& \cellcolor{gg}\textbf{66.2} 
& \cellcolor{gg}59.9 
& \cellcolor{gg}\textbf{36.3} 
& \textbf{64.9} 
& \textbf{98.1\%} \\
    \rowcolor{ggr}\multicolumn{14}{c}{\textit{Keep 32 Tokens} \textcolor{Green}{(\textbf{Pruning Ratio} $\mathbf{94.4\%}$)}} \\
    PruMerge+ &ICCV-2025 & 66.8 & 54.1 & 52.3 & 71.7 & 52.4 & 67.4 & 1269.1 & 61.1 & 53.5 & 28.7 & 57.1 & 86.5\% \\
    TRIM &COLING-2025 & 69.8 & 55.6 & 48.8 & 70.4 & 49.6 & 85.8 & 1284.7 & 63.1 & 45.4 & 26.4 & 57.9 & 86.4\% \\
    VisionZip &CVPR-2025 & 68.4 & 52.7 & 53.0 & 72.9 & 55.2 & 66.8 & 1257.7 & 61.2 & 55.8 & 29.3 & 57.8 & 87.6\% \\
    DART &EMNLP-2025 & 68.1 & 53.9 & 52.0 & \textbf{73.2} & 55.1 & 66.9 & 1282.8 & 61.9 & 56.2 & 29.4 & 58.1 & 88.0\% \\
    DivPrune &CVPR-2025 & 72.0 & 56.2 & \textbf{54.5} & 70.9 & 54.6 & 79.3 & 1405.2 & 61.7 & 57.2 & 27.8 & 60.4 & 90.8\% \\
    CDPruner &NeurIPS-2025 & 75.2 & 58.5 & 53.5 & 71.9 & 55.3 & 87.6 & 1421.0 & 63.7 & 56.6 & 30.9 & 62.4 & 93.8\% \\
\cellcolor{gg}\textbf{CAPruner}{ \small(Ours)} &\cellcolor{gg}This paper 
& \cellcolor{gg}\textbf{76.6} 
& \cellcolor{gg}\textbf{59.4} 
& \cellcolor{gg}\textbf{54.5} 
& \cellcolor{gg}72.6 
& \cellcolor{gg}\textbf{57.0} 
& \cellcolor{gg}\textbf{88.2} 
& \cellcolor{gg}\textbf{1460.0} 
& \cellcolor{gg}\textbf{64.5} 
& \cellcolor{gg}\textbf{58.0} 
& \cellcolor{gg}\textbf{33.4} 
& \textbf{63.7} 
& \textbf{96.0\%} \\
    \bottomrule
    \end{tabular}
  }
\end{table*}
\label{add-LLaVA-1.5-13B}
\begin{table*}[t]
  \centering
    \caption{Main results on InternVL3-8B under different visual
    token budgets.
    \textbf{Acc.} denotes the average performance across
    diverse benchmarks, and \textbf{Rel.} denotes the mean
    of per-benchmark performance ratios relative to the
    full-token baseline.
    The best results are highlighted in \textbf{bold}.}
  \label{tab:internvl3}
  \resizebox{\linewidth}{!}{
    \begin{tabular}{l|c|cccccccc|aa}
    \toprule
    \textbf{Method} &\textbf{Venue}  & \textbf{AI2D} & \textbf{TextVQA} & \textbf{ChartQA} & \textbf{OCRBench} & \textbf{HallBench} & \textbf{MME} & \textbf{MMB-EN} & \textbf{MMB-CN} & \textbf{Acc.} & \textbf{Rel.} \\
    \rowcolor{ggr}\multicolumn{12}{c}{\textit{Full-token Baseline, All 1280 Tokens} ($\mathbf{100\%}$)} \\
    InternVL3-8B &- & 85.2 & 81.5 & 85.1 & 853 & 50.0 & 2394 & 83.9 & 82.6 & 84.2 & 100.0\% \\
    \rowcolor{ggr}\multicolumn{12}{c}{\textit{Keep 256 Tokens} \textcolor{Green}{(\textbf{Pruning Ratio} $\mathbf{80.0\%}$)}} \\
    FastV &ECCV-2024 & 82.2 & 74.4 & 70.7 & 632 & 48.5 & 2348 & 83.6 & 82.0 & 77.8 &  92.4\% \\
    DivPrune &CVPR-2025 & 80.9 & 64.7 & 57.5 & 477 & 38.7 & 2249 & 80.8 & 80.2 & 70.4 &  82.8\% \\
    CDPruner &NeurIPS-2025 & 82.7 & 75.7 & 72.0 & 640 & 48.8 & 2334 & 83.5 & 81.7 & 78.1 &  92.9\% \\
\cellcolor{gg}\textbf{CAPruner}{ \small(Ours)} & \cellcolor{gg}This paper
& \cellcolor{gg}\textbf{83.7} 
& \cellcolor{gg}\textbf{76.6} 
& \cellcolor{gg}\textbf{73.0} 
& \cellcolor{gg}\textbf{666} 
& \cellcolor{gg}\textbf{49.4} 
& \cellcolor{gg}\textbf{2366} 
& \cellcolor{gg}\textbf{84.0} 
& \cellcolor{gg}\textbf{82.6} 
& \textbf{79.3} 
& \textbf{94.2\%} \\
    \rowcolor{ggr}\multicolumn{12}{c}{\textit{Keep 128 Tokens} \textcolor{Green}{(\textbf{Pruning Ratio} $\mathbf{90.0\%}$)}} \\
    FastV &ECCV-2024 & 77.3 & 63.7 & 46.9 & 426 & 42.5 & 2250 & 81.3 & 80.2 & 68.4 &  80.9\% \\
    DivPrune &CVPR-2025 & 76.4 & 55.6 & 42.7 & 378 & 37.7 & 2166 & 78.4 & 77.6 & 64.3 &  75.7\% \\
    CDPruner &NeurIPS-2025 & 79.9 & 67.5 & 50.8 & 471 & 44.6 & 2282 & 82.1 & 80.3 & 70.8 &  83.9\% \\
\cellcolor{gg}\textbf{CAPruner}{ \small(Ours)} &\cellcolor{gg}This paper
& \cellcolor{gg}\textbf{81.0} 
& \cellcolor{gg}\textbf{69.4} 
& \cellcolor{gg}\textbf{52.7} 
& \cellcolor{gg}\textbf{504} 
& \cellcolor{gg}\textbf{45.4} 
& \cellcolor{gg}\textbf{2310} 
& \cellcolor{gg}\textbf{83.0} 
& \cellcolor{gg}\textbf{81.5} 
& \textbf{72.4} 
& \textbf{85.8\%} \\
    \bottomrule
    \end{tabular}
  }
\vspace{-3mm}
\end{table*}
\label{add-InternVL3-8B}
\section{Details of Experimental Setup}
\label{details}
All experiments are conducted on eight NVIDIA RTX 4090 GPUs with a batch size of 1. CAPruner is fully training-free and does not require additional model training, task-specific fine-tuning, or modification of the pretrained backbone weights. The visual--text association coefficients are estimated on an independent calibration set and reused during inference. We evaluate CAPruner on five model configurations spanning four representative MLLM architectures with distinct visual tokenization designs. Qwen2.5-VL-7B~\cite{bai2025qwen25vltechnicalreport} employs a dynamic-resolution ViT encoder with window attention, SwiGLU activation, and RMSNorm, producing 1,296 visual tokens under the evaluated 1008$\times$1008 input resolution. LLaVA-1.5-7B/13B~\cite{liu2024improved} adopt a CLIP-based visual encoder~\cite{radford2021learning} and an MLP projector to bridge visual features with a Vicuna language model~\cite{vicuna2023}, producing 576 visual tokens from 336$\times$336 image inputs. LLaVA-NeXT-7B extends the LLaVA architecture to high-resolution inputs through dynamic image patching, producing up to 2,880 visual tokens in our evaluation. InternVL3-8B~\cite{zhu2025internvl3} adopts a ViT-MLP-LLM architecture with an MLP connector, native multimodal pre-training, and variable visual position encoding, producing 1,280 visual tokens under the evaluated 896$\times$896 input resolution.

For pruning configurations, we retain 128, 64, and 32 visual tokens for LLaVA-1.5-7B/13B, corresponding to pruning ratios of 77.8\%, 88.9\%, and 94.4\%, respectively. For Qwen2.5-VL-7B, we retain 512, 256, and 128 tokens, corresponding to pruning ratios of 60.5\%, 80.2\%, and 90.1\%, respectively. For InternVL3-8B, we retain 256 and 128 tokens, corresponding to pruning ratios of 80\% and 90\%, respectively. For LLaVA-NeXT-7B, the end-to-end efficiency evaluation uses a maximum budget of 320 retained visual tokens. For aggregate performance evaluation, Acc. is computed
as the mean benchmark score after scaling MME by $1/20$
and OCRBench by $1/10$, where applicable.
Rel. is computed as the mean of per-benchmark performance
ratios relative to the corresponding full-token baseline.

\section{Additional Experimental Results}
This section extends the main-paper evaluation with additional results on LLaVA-1.5-7B, LLaVA-1.5-13B, and InternVL3-8B, as well as video understanding benchmarks using Qwen2.5-VL-7B.
\subsection{Comparisons on LLaVA-1.5-7B}
\label{add-LLaVA-1.5-7B}
As shown in Table~\ref{tab:llava-1.5}, CAPruner achieves the best overall performance across all pruning ratios. With 128 retained tokens, it preserves 100.4\% of the original performance, reaching an average score of 63.3 and nearly matching the full-token baseline. When the budget is reduced to 64 tokens, several pruning baselines exhibit noticeable performance degradation, while CAPruner retains 98.4\% relative performance, outperforming CDPruner and DivPrune by 1.4 and 3.7 percentage points, respectively. Under the most aggressive 32-token setting, CAPruner preserves 96.1\% of the original performance and exceeds CDPruner by 1.1 average-score points. It also maintains strong POPE performance across all budgets, even surpassing the unpruned baseline, suggesting that CAPruner can preserve task-relevant visual information under aggressive token reduction.
\begin{table*}[t]
\centering
\small
\caption{Evaluation of visual token pruning on video
understanding benchmarks using Qwen2.5-VL-7B
under a 75\% pruning ratio.
\textbf{Rel.} denotes the mean of per-benchmark
performance ratios relative to the unpruned backbone.}
\label{tab:video}
\resizebox{0.95\linewidth}{!}{
\begin{tabular}{l|c|ccca}
\toprule
\textbf{Method} 
& \textbf{Venue}
& \textbf{Vinoground (Group)}
& \textbf{VideoMME (Perception)} 
& \textbf{SEED-Bench (All)} 
& \textbf{Rel.} 
\\
\midrule
Qwen2.5-VL-7B (baseline) &- & 20.20 & 61.33  & 74.20 & 100.0\% \\
FastV &ECCV-2024 & 12.80 & 59.44  & 69.22 & 84.52\% \\
VisionZip &CVPR-2025 & 12.80 & \textbf{59.67} & 72.11 & 85.95\% \\
DivPrune &CVPR-2025 & 14.00 & 58.00 & 72.11 & 87.02\% \\
DART &EMNLP-2025 & 12.60 & 59.52  & 71.27 & 85.16\% \\
IDPruner &arXiv-2026 & 13.40 & 59.48 & 72.68 & 87.09\% \\
\cellcolor{gg}\textbf{CAPruner} {\small(Ours)}
&\cellcolor{gg}This paper
& \cellcolor{gg}\textbf{14.60}  
& \cellcolor{gg}59.50 
& \cellcolor{gg}\textbf{72.90}  
& \textbf{89.18\%} \\
\bottomrule
\end{tabular}
}
\vspace{-1mm}
\end{table*}
\subsection{Comparisons on LLaVA-1.5-13B}
To verify whether CAPruner scales to larger MLLM backbones, we conduct additional experiments on LLaVA-1.5-13B under the same token budgets. As shown in Table~\ref{tab:LLaVA-1.5-13B}, CAPruner also achieves the best overall performance on the larger LLaVA-1.5-13B backbone across all pruning ratios. With 128 retained tokens, it preserves 99.0\% of the original performance, reaching an average score of 65.5 and outperforming CDPruner by 0.7 average points. When the budget is reduced to 64 tokens, CAPruner retains 98.1\% relative performance, exceeding CDPruner and DivPrune by 1.5 and 4.0 percentage points, respectively. Under the most aggressive 32-token setting, CAPruner still preserves 96.0\% of the full-token performance and improves over CDPruner by 1.3 average points. Notably, CAPruner consistently maintains strong POPE performance across all budgets, even surpassing the unpruned baseline, suggesting that sensitivity-guided token selection remains effective on larger MLLM backbones.
\subsection{Comparisons on InternVL3-8B}

Beyond the LLaVA-family architectures, we further evaluate CAPruner on InternVL3-8B, an advanced open-source MLLM with a ViT-MLP-LLM architecture and extended multimodal contexts. As shown in Table~\ref{tab:internvl3}, CAPruner consistently achieves the best average performance under both pruning ratios. With 256 retained tokens, it preserves 94.2\% of the full-token performance, reaching an average score of 79.3 and outperforming CDPruner by 1.2 average points. When the budget is further reduced to 128 tokens, CAPruner still retains 85.8\% relative performance and improves over CDPruner by 1.6 average points. It also achieves consistent gains on text-oriented and reasoning-intensive benchmarks such as TextVQA, ChartQA, OCRBench, MME, and MMBench. These results suggest that the proposed sensitivity-guided token selection can generalize beyond the LLaVA family to MLLM architectures with different visual encoding and alignment designs.
\subsection{Generalization to Video Understanding}
\label{video}
Although CAPruner is primarily developed for image-based MLLMs, its intervention-oriented token importance estimation can also be applied to video-language models. We evaluate this generalization ability on Qwen2.5-VL-7B using three representative video benchmarks: Vinoground~\cite{zhang2024vinoground}, VideoMME~\cite{fu2025video}, and SEED-Bench~\cite{li2024seed}. Vinoground evaluates fine-grained video-text alignment with the Group metric, VideoMME measures video perception ability with the Perception subset, and SEED-Bench reports overall video understanding performance with the All metric. For video evaluation, we uniformly sample 32 frames
per video and employ the native dynamic-resolution
preprocessing of Qwen2.5-VL-7B.
Given $N_v$ visual tokens across the sampled frames,
we globally retain $\lfloor 0.25N_v \rfloor$ tokens
without imposing frame-wise retention quotas, while
preserving their original spatiotemporal position indices.
CAPruner follows the same pruning layers, hyperparameters,
and calibration coefficients as in the image evaluation.
All compared methods use identical video inputs,
preprocessing, and final visual-token budgets. Table~\ref{tab:video} summarizes the comparison with existing visual token pruning methods. CAPruner achieves the best average relative performance among all pruning methods, reaching 89.18\% under a 75\% pruning ratio. Compared with IDPruner, CAPruner improves the average relative performance by 2.09 percentage points. These results suggest that the proposed sensitivity-guided token selection can extend beyond image-based MLLMs while preserving useful visual information for video understanding.

\end{document}